\documentclass[10pt,twocolumn,letterpaper]{article}

\usepackage{iccv}
\usepackage{times}
\usepackage{epsfig}
\usepackage{graphicx}
\usepackage{amsmath}
\usepackage{amssymb}

\usepackage[breaklinks=true,bookmarks=false]{hyperref}

\iccvfinalcopy 

\def\iccvPaperID{****} 

\ificcvfinal\fi

\begin{document}

\title{Attribute-Driven Spontaneous Motion in Unpaired Image Translation}

\author{Ruizheng Wu$^{1}$ \quad Xin Tao$^{2}$ \quad Xiaodong Gu$^{3}$ \quad Xiaoyong Shen $^{2}$ \quad Jiaya Jia$^{1,2}$\\
	\normalsize $^{1}$The Chinese University of Hong Kong \quad $^{2}$Tencent YouTu Lab \quad $^{3}$ Harbin Institute of Technology, Shenzhen\\
	{\tt\small \{rzwu, leojia\}@cse.cuhk.edu.hk} \quad \tt\small \{xintao, dylanshen\}@tencent.com \tt\small guxiaodong@stu.hit.edu.cn}

\maketitle

\begin{abstract}
	Current image translation methods, albeit effective to produce high-quality results in various applications, still do not consider much geometric transform. We in this paper propose the spontaneous motion estimation module, along with a refinement part, to learn attribute-driven deformation between source and target domains. Extensive experiments and visualization demonstrate effectiveness of these modules. We achieve promising results in unpaired-image translation tasks, and enable interesting applications based on spontaneous motion.
\end{abstract}

\let\thefootnote\relax\footnotetext{Project page: ~ \href{https://github.com/mikirui/ADSPM}{https://github.com/mikirui/ADSPM}}

\section{Introduction}\label{sec:intro}
High-quality image generation is a fascinating task and has gained much attention in computer vision community. There has been great progress using generative adversarial networks (GAN)~\cite{goodfellow2014generative, radford2015unsupervised}. \emph{Image translation}, which produces modified images in target domain based on a given input from source domain, has been widely used in applications of style transfer~\cite{li2016precomputed}, sketch/photo conversion~\cite{chen2018sketchygan, pix2pix2017}, label-based image synthesis~\cite{qi2018semi}, face editing~\cite{xiao2018elegant}~\etc. Recent research trends continuously towards high practicality, e.g., images in high resolutions~\cite{wang2018high}, and unpaired image
translation~\cite{zhu2017unpaired}, or using better latent space for more effective control~\cite{huang2018multimodal, lee2018diverse}.

\def\widtht{0.31}
\begin{figure}[h]
	\begin{center}
		\begin{tabular}
			{@{\hspace{0.0mm}}c@{\hspace{2mm}}c@{\hspace{1mm}}c@{\hspace{0.0mm}}}
			\includegraphics[width=\widtht\linewidth]{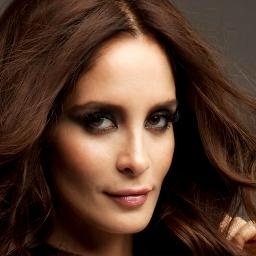}&
			\includegraphics[width=\widtht\linewidth]{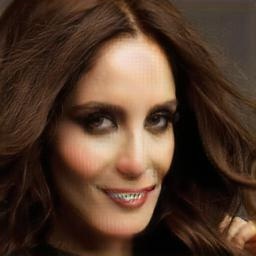}&
			\includegraphics[width=\widtht\linewidth]{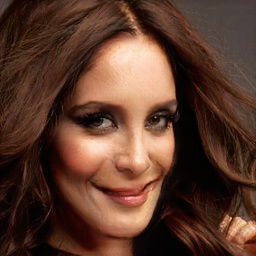}\\
			\small	(a) Input &  \small (b) StarGAN~\cite{choi2018stargan} & \small (c) CycleGAN~\cite{zhu2017unpaired} \\
			\includegraphics[width=\widtht\linewidth]{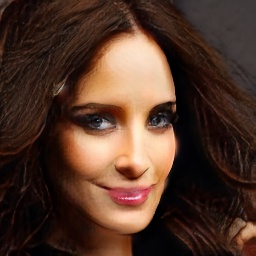}&
			\includegraphics[width=\widtht\linewidth]{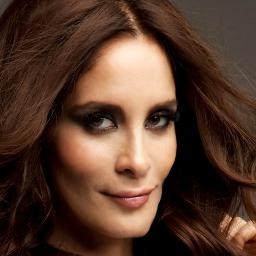}&
			\includegraphics[width=\widtht\linewidth]{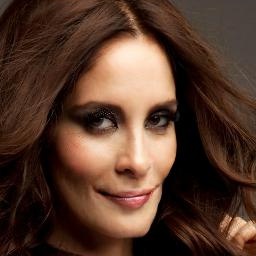}\\
			\small (d) MUNIT~\cite{gokaslan2018improving} & \small (e) Ours (SPM) & \small (f) Ours (SPM + R) \\
		\end{tabular}
	\end{center}
	\vspace{-0.05in}
	\caption{Examples of nonsmile-to-smile faces transform, where ``Ours (SPM)" shows our deformed result and ``Ours (SPM + R)" indicates our final result after further refinement.}
	\vspace{-0.15in}
	\label{fig:compare}
\end{figure}

Image translation mostly imposes the requirement of aligned or similar domains for texture or appearance transform. For example, in style transfer, the output image generally shares the same content with input.
The building blocks of these networks, such as convolution/deconvolution layers and activation functions, are spatially corresponding. As shown in Fig.~\ref{fig:compare}(a)-(d), visual artifacts, such as ghosting, could appear when nonsmile and smile faces are not geometrically aligned in image space.

In this paper, we take advantage of geometric correspondence in appearance transform. Taking the smiling face as an example in Fig.~\ref{fig:compare}(e)-(f), decent results can be produced by applying geometric transform only, and they are better when a small refinement network follows. This architecture can greatly reduce visual artifacts mentioned above. This paper tackles the following three issues.

\vspace{-0.2in}
\paragraph{Single Image Deformation} Although several previous methods~\cite{yin2019geogan, yeh2016semantic, tao2017detail} employ various flow/warping/deformation by estimating
motion, they are different from what we need in two aspects. First, traditional motion field estimation requires a pair of images to construct dense pixel correspondence, while in our task only one image is available. Second, motion fields for
deformation are conditioned by examples, where one input image may need various motion constrained by target examples. 

Different from all these settings, our goal is more like estimation of natural tendency of input images. We term it \emph{spontaneous motion} (SPM) to distinguish from ordinary optical flow. This new tool adds a new dimension to image translation by introducing unpaired geometric transform. It also enables new ways of visualization, and finds interesting applications (described in Sec. \ref{sec:analysis}). For example, in our framework, SPM for different target domains can be viewed as motion basis (Fig.~\ref{fig:warp_field}), and linearly combining SPM basis enables convenient geometric edit (Fig.~\ref{fig:flow_comb}).


\vspace{-0.15in}	
\paragraph{High Ill-posedness} Our framework is trained with neither paired data nor ground-truth motion field across domains.
Cycle reconstruction loss and learning common latent space were considered to deal with unpaired data~\cite{zhu2017unpaired, liu2017unsupervised}. Our geometric transform estimation across domains is even more ill-posed, since this set-to-set motion is more ambiguous compared to image-to-image correspondence given no ground-truth motion. Our spontaneous motion module applies two domain classifiers for translation result generation and motion estimation.

\vspace{-0.15in}
\paragraph{Inevitable Errors}
Estimated motion fields are inevitably with errors due to large prediction freedom, missing-motion area to be filled (teeth of smiling faces are area missing in non-smile faces), and fine texture requirement for high quality results. Our system has a \emph{refinement} module to fix remaining visual artifacts with an attention mask to filter out unnecessary changes on original images.

Our contributions are as follows.
1) We propose an end-to-end unpaired image translation system considering geometric deformation.
2) A conditional spontaneous motion estimation module, along with domain classifiers and a refinement step, to boost performance.
3) Our new framework achieves promising results in image translation, especially for unaligned scenarios.

\section{Related Work}
\paragraph{Unpaired Image Translation}
Several unsupervised image translation methods were proposed. By introducing cycle-consistency loss for reconstruction, methods of \cite{zhu2017unpaired,yi2017dualgan,kim2017learning} train the translation network across two domains without paired data.
They train two separate networks for bidirectional image generation between source and target domains.
Methods of \cite{choi2018stargan, pumarola2018ganimation} extend the framework by introducing additional conditions to generate images in multiple domains. Another stream of research~\cite{liu2016coupled,liu2017unsupervised,taigman2016unsupervised,wolf2017unsupervised,murez2017image} is based
on the assumption that images in source and target domains share the same latent space and in \cite{huang2018multimodal, lee2018diverse}, style and content are disentangled to control generated image style. They successfully translate images across domains. Because geometric relationship between domains is not considered, data that is not aligned or structurally very different cannot be well dealt with.

\vspace{-0.15in}
\paragraph{Geometry-Aware Image Translation}
There exists work to build geometric relationship during image translation/generation. In \cite{gokaslan2018improving,mechrez2018contextual}, geometric inconsistency between domains is mitigated with designed discriminator or losses. We note the generators are still composed of convolution-based blocks, which limit the generation power.
Methods of \cite{dong2018soft, yin2019geogan, geng2018warp, yeh2016semantic} estimate correspondence between two images. 
Dong et al. \cite{dong2018soft} relied on human body parsing, while Geng et al. \cite{geng2018warp} generated dense correspondence based on face landmarks. Methods of \cite{yin2019geogan, yeh2016semantic} directly learn dense 
correspondence between two images. For these methods, paired reference images are needed to train or test, which does not fit semantic-level set-to-set transformation. Cao et al. \cite{cao2018cari} added another network for landmark learning; it cannot be trained in an end-to-end manner.

Our method is different. We do not need reference images and our framework is designed in an end-to-end way. 
Besides, the estimated spontaneous motion is conditioned on source domain content and target domain attributes, which can achieve semantic-level geometric transformation and generation.

\section{Proposed Method}\label{sec:method}
\begin{figure*}
	\begin{center}
		\includegraphics[width=\linewidth]{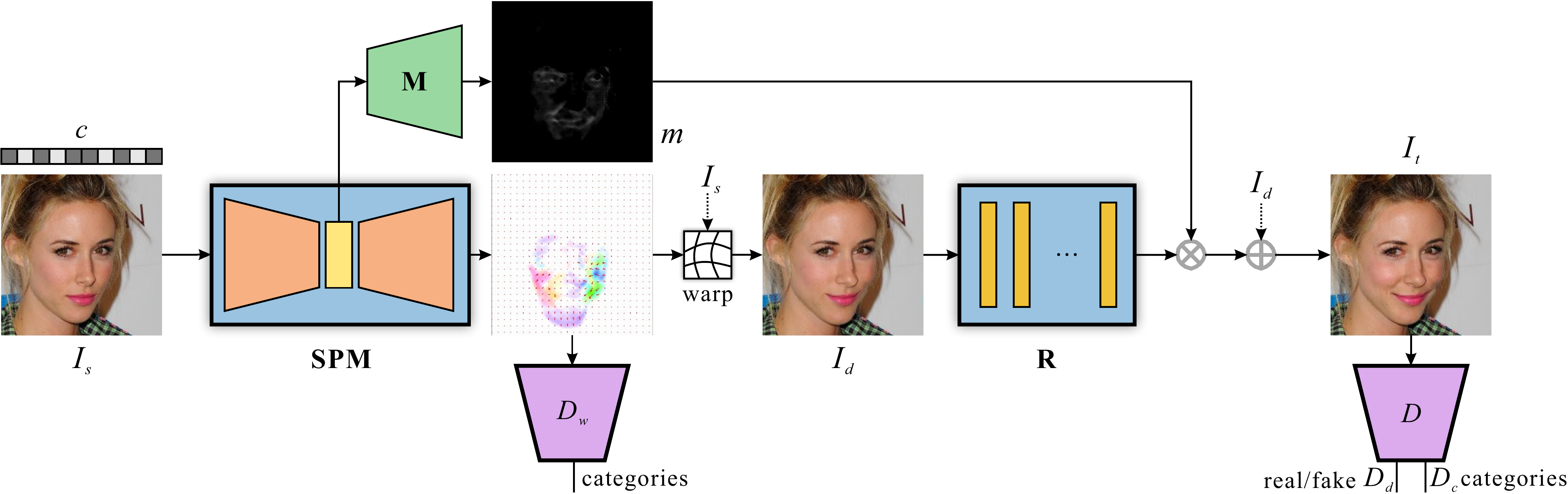}
	\end{center}
	\vspace{-0.1in}
	\caption{Our overall framework. Our generator $G$ contains spontaneous motion module $SPM$ and refinement module $R$. Two domain classifiers $D_w$ and $D_c$ are utilized to drive generation of final results and motion fields under different conditions, while $D_d$ is utilized to distinguish real images from fake ones.}
	\label{fig:overall}
\end{figure*}

Given an image in source domain $I_s \in \mathbb{R}^{H \times W \times 3}$ and target domain indicator $c_t \in \{0,1\}^{N}$ ($N$ is the total number of attributes), \eg~smiling, angry, and surprising. Our goal is to generate a high-quality image $I_t$ with attribute $c_t$ while keeping the identity of $I_s$.
Our framework resembles previous generative models by iteratively training generator/discriminator networks. However, in order to better handle geometric transform, we incorporate two new modules in
generator $\mathbf{G}$ as spontaneous motion module $\mathbf{SPM}$ (Sec.~\ref{sec:warpfield}) and refinement module $\mathbf{R}$ (Sec.~\ref{sec:refine}). Two types of classifiers are proposed as new
losses to facilitate training. We extend our framework to high-resolution image generation ($512\times512$) with special designs (Sec. \ref{sec:hires}).	
Our overall framework is depicted in Fig.~\ref{fig:overall}. We elaborate on each module in the following.

\vspace{-0.05in}

\subsection{Spontaneous Motion Module}\label{sec:warpfield}
According to the analysis in Sec.~\ref{sec:intro}, our spontaneous motion module aims to predict motion field $w$ based on input image $I_s$ and target indicator $c_t$. We use an encoder-decoder network structure for its powerful fitting ability. For the design of this module, we consider the following facts.

\vspace{-0.1in}	
\paragraph{Motion Field Decoding} 
In conventional image regression problems \cite{ilg2017flownet, tao2017detail, dong2014learning}, activation functions in the final layer is usually not applied, in order to leave the output unbounded. This is because ground-truth pixel values
are always in range $[0,1]^{H\times W\times 3}$, which supervise and prevent network output from divergence.
In our motion estimation task, contrarily the output motion values can be largely varied,
and the network can only be trained under indirect supervision, making convergence an issue, as verified in our experiments.

In order to mitigate this problem, we utilize $\textbf{tanh}()$ as the last activation function to limit output in $[-1, 1]$ rather than $[-\infty, +\infty]$.
Moreover, we introduce an empirical multiplier $\lambda_{w}$ to get the final $w$. This seemingly tricky coefficient is actually quite reasonable in many tasks,
such as face editing, since only local deformation is needed.
In our paper, we set $\lambda_{w}$ to $[0.1, 0.2]$ for different datasets. Visualization of our estimated motion field with different targets
is given in Fig. \ref{fig:warp_field}. 	

\vspace{-0.1in}	
\paragraph{Motion Field Direction}\label{sec:warping}
To deform input to the target image, backward motion is usually considered as a vector from the target to source points~\cite{ilg2017flownet, dong2018soft, siarohin2018animating}. 
However, such motion representation may not be suitable for convolution/deconvolution layer networks with aligned operators, since the representation is aligned to the unknown deformed image rather than the input one. Forward motion can mitigate the problem to some extent due to the alignment between input and forward motion. But deforming with forward warping may bring holes and more artifacts than with the backward one. 

We experiment with these two solutions and adopt backward motion representation for deforming, because its result contains less artifacts and is more stable.
Denoting image coordinates as $i = (x, y)^{T}$, the set of valid image coordinates as $V$, input image as $I_s(i)$, deformed image as $I_d(i)$ and motion field as $w(i) = (u(i), v(i))^{T}$, we formulate the deformation step with bilinear interpolation as
\begin{equation}\label{eq: calc_w}
I_d(i) =
\left\{
\begin{array} {ll}
\widetilde{I_s}(i + w(i)), & \text{if $i + w(i)$ $\in$ $V$}, \\
0, & \text{otherwise}.
\end{array}
\right.
\end{equation}
where $\widetilde{I_s}(i)$ is bilinear interpolation operator.

As for the network structure of spontaneous motion module, we construct the encoder with 3 stride-2 convolution layers (each followed by instance normalization and ReLU) and 6 residual blocks to extract $8\times$ down-sampled feature map $f$. A decoder then processes and up-samples $f$ by 3 deconvolution layers to a 2-channel motion field with the same size as the input image. In addition, high-level feature $f$ is utilized for attention mask learning, which will be described in Sec. \ref{sec:refineatt}.

To generate different dense motion fields for target indicators, we design two classifiers as constraints for both generation results and motion field estimation.

\vspace{-0.1in}	
\paragraph{Domain Classifiers}
We design image and motion domain classifiers in training. For image classifier $D_{c}$, like that of \cite{choi2018stargan}, we add $D_{c}$ on top of discriminator $D$ as a constraint to classify the generated images into target domain $c$. 
During training on $D$, real image $I_s$ and its attribute $c_{s}$ are utilized to train $D_{c}$ with loss $\mathcal{L}_{cr}^{d}$.
At the stage of training generator $G$, $D_{c}$ is fixed and the classification loss $\mathcal{L}_{cr}^{g}$ of generated images is utilized to optimize $G$.
The losses $\mathcal{L}_{cr}^{d}$ and $\mathcal{L}_{cr}^{g}$ are defined as
\begin{equation}\label{eq:real_res}
\mathcal{L}_{cr}^{d} = \mathbb{E}_{I_{s}, c_{s}}[-log D_{c}(c_{s} | I_{s})]
\end{equation}
\vspace{-0.15in}
\begin{equation}\label{eq:gen_res}
\mathcal{L}_{cr}^{g} = \mathbb{E}_{I_{s}, c_{t}}[-log D_{c}(c_{t} | G(I_{s}, c_{t}))]
\end{equation}
Although the classifier for generation results can guide the prediction of motion, we note a constraint on motion helps it more directly and better -- motion fields for one domain (such as a face expression exemplified later in Fig. \ref{fig:warp_field}) have common features. It does not vary much even with different input images.
With this observation, we design a classifier $D_{w}$ for motion fields, which classifies different motion fields into categories according to the target condition. It makes motion under different conditions share similar patterns and thus reduces bias or noise in generation steps.
The classification loss for this classifier is formulated as
\begin{equation}\label{eq:gen_warp}
\mathcal{L}_{cw} = \mathbb{E}_{I_s, c_{t}}[-log D_{w}(c_{t} | SPM(I_{s}, c_{t}), I_s)].
\end{equation}

\subsection{Refinement Module}\label{sec:refine}
The deformed image $I_d$ is further refined to reduce artifacts and enhance textural details. Specifically, two components are used.

\vspace{-0.1in}		
\paragraph{Refinement with Residual Learning}\label{sec:refineres}
We employ a refinement sub-network after deforming image $I_d$. Instead of directly learning in images space, we learn the residual $r$ between deformed image and the unknown target, \ie $I_{t} = I_d + r$, since to learn residual for a well deformed image is easier and more reliable.
As for the network structure, $n$ residual blocks~\cite{he2016deep} are sequentially concatenated without any downsample operation. The residual blocks are used for finer structure update. Thus we do not shrink images spatially, and instead take multiple stacked residual blocks to ensure final effect. In our experiments, we set $n=12$ to balance 
performance and efficiency.

\vspace{-0.1in}	
\paragraph{Attention Mask}\label{sec:refineatt}
In image translation, generally only essential regions need to be updated (e.g. only mouth and its surrounding are changed when transforming neutral faces to smiling ones). We propose learning an attention mask $m$, which marks important regions. The results are denoted as $I_{t} = I_d + r \cdot m$.

Specifically, as mentioned in Sec. \ref{sec:warpfield}, we obtain down-sampled feature map $f$ in module $SPM$. $f$ catches high-level semantic information. We utilize it for attention mask learning. We build the attention mask module $M$ with 3 deconvolutional layers to up-sample $f$ into a 1-channel mask $m$, the same size as the input. $Sigmoid$ layer is used as the final activation layer to range the output mask in $[0, 1]$.

Directly learning an attention mask without any additional constraints is difficult, due to possible trivial solution of a mask with all-region selected. To avoid this problem, we introduce a regularization term $\mathcal{L}_{m}$ to enforce sparsity of masks in $\mathcal{L}_1$-norm:
\begin{equation}\label{eq:mask}
\mathcal{L}_{m} = \frac{1}{CHW} \sum_{i=1}^{C} \sum_{j=1}^{H} \sum_{k=1}^{W} |m_{ijk}|,
\end{equation}
where $C$, $H$ and $W$ are channel number, height and width of the mask respectively. The loss forces the attention mask to focus on the most important region.

\subsection{Higher Resolution}\label{sec:hires}
To generate high resolution (HR) image by a single generator is difficult. Previous work \cite{karras2017progressive, li2018unsupervised, karras2018style} adopted coarse-to-fine 
or multi-stage training strategies. By incorporating motion estimation and refinement modules, we extend previous 
coarse-to-fine strategies to a pipeline with extra priors.

\vspace{-0.1in}		
\paragraph{Priors and Adaptation}
Previous coarse-to-fine strategies are usually applied to final output, \ie using generated low-res (LR) images to guide HR generation. In our framework, we have more useful clues from LR, \ie motion field $w^l$ for deformation, residual $r^l$ for refinement, mask $m^l$ for attention in low-resolution form. We utilize them to facilitate HR result generation.

We first train our initial framework with LR images ${I_{s}^l}$ until convergence. For higher resolution results, we feed in HR image ${I_{s}^h}$ and start from the well-trained LR framework while the weight of LR model is updated simultaneously in this stage.	
After obtaining motion field $w^l$, residual $r^l$, attention mask $m^l$ from LR generator with down-sampled $I_s^l$ from $I_s^h$, we up-sample them to produce coarse results, \ie $U(w^l)$, $U(r^l)$ and $U(m^l)$ with the same size as HR images ${I_{s}^h}$. We further incorporate three light-weighted enhancement networks ($T_w$, $T_r$ and $T_m$) respectively, each only contains two convolutional layers and a
residual block. Finally, we estimate motion field $w^h$ as
\begin{equation}\label{eq:hires_warp}
w^h = U(w^l) + T_w(U(w^l)).
\end{equation}
Bilinear upsampling is used with the same process to obtain residual $r^h$ and attention mask $m^h$. With these intermediate results, we deform $I_{s}^h$ by $w^h$ to get $I_d^{h}$ and then refine $I_d^{h}$ to yield final output $I_{t}^h = I_{d}^h + r^h * m^h$.

\vspace{-0.1in}	
\paragraph{Resolution Adaptive Discriminator}
During training, the discriminators are designed as follows. In the LR-image training stage, we only train the LR image discriminator $D^l$. We set real image $I_{s}^l$ as the positive sample while the generated $I_{t}^l$ is the negative one. 
In the HR image training stage, for $D^l$, we have down-sampled $I_{s}^l$ as the positive sample and generated LR image $I_{t}^l$ as the negative one. Besides, we down-sample generated HR image $I_t^h$ to LR and feed them to $D^l$ as another type of negative samples. As for $D^h$, $I_s^h$ and final generation result $I_t^h$ are positive and negative samples respectively. $D^h$ share similar network structure as $D^l$, and yet with more convolution layers.

\subsection{Other Loss Functions}
\paragraph{Adversarial Loss}
Ordinary generative adversarial loss is set for $G$ and $D_d$ formulated as
\begin{equation}\label{eq:gan_loss}
\begin{split}
\mathcal{L}_{adv} = &\mathbb{E}_{I_s}[log D_d(I_s)] + \\
&\mathbb{E}_{I_s,c_t}[log(1 - D_d(G(I_s, c_t)))].
\end{split}
\end{equation}
\paragraph{Reconstruction Loss}
Similar to \cite{choi2018stargan, zhu2017unpaired}, we reconstruct images in cycle flow. With source image $I_s$, generated image $I_t$, and source image attribute $c_{s}$, we formulate the reconstruction loss $\mathcal{L}_{rec}$ as
\begin{equation}\label{eq:rec_loss}
\mathcal{L}_{rec} = \mathbb{E}_{I_s, c_t, c_s}\|I_s - G(G(I_s, c_t), c_s)\|_{1}.
\end{equation}
\paragraph{Total Loss}
The final loss function for generator $G$ is
\begin{equation}\label{eq:total_g}
\begin{split}
\mathcal{L}_{g} = &\lambda_{cr} \cdot \mathcal{L}_{cr}^{g} + \lambda_{cw} \cdot \mathcal{L}_{cw} + \\
&\lambda_{m} \cdot \mathcal{L}_{m} + \lambda_{adv} \cdot \mathcal{L}_{adv} + \lambda_{rec} \cdot \mathcal{L}_{rec}.
\end{split}
\end{equation}
The loss function for $D$ is
\begin{equation}\label{eq:total_d}
\mathcal{L}_{d} = \lambda_{cr} \cdot \mathcal{L}_{cr}^{d} - \lambda_{adv} \cdot \mathcal{L}_{adv}.
\end{equation}
In our experiments, we set $\lambda_{cr}$, $\lambda_{m}$, and $\lambda_{adv}$ all to 1.0, and set $\lambda_{rec}$ and $\lambda_{cw}$ to 10.0 and 0.5 respectively.

\section{Experiments}
We conduct experiments on both CelebA \cite{liu2015faceattributes} and RaFD \cite{langner2010presentation}. CelebA contains 200K celebrity images and 40 attributes for each image with resolution $218 \times 178$. We utilize CelebA-HQ \cite{karras2017progressive} in resolution $1024 \times 1024$ for high-res image usage. To demonstrate the effectiveness of our framework, we select attributes with geometric deformation, \ie `Smiling', `Arched\_eyebrow', `Big\_Nose', and `Pointy\_nose' as condition to train our framework. RaFD is a smaller dataset with 67 identities, each displaying 8 emotional expressions, 3 gaze directions and 5 camera angles. We only train on frontal faces for robustness.

We implement the system on PyTorch \cite{paszke2017automatic} and run it on a TITAN Xp card. During our two-stage training, we first train on LR framework with $128 \times 128$ images and batch size 16 for $1 \times 10^5$ iterations. Then we train our extended network on higher resolutions $256 \times 256$ or $512 \times 512$ with batch size 8 for another $2 \times 10^5$ iterations. We use Adam \cite{kingma2014adam} with learning rate 1e-4 to optimize our framework.

\subsection{Analysis}\label{sec:analysis}
\paragraph{Effectiveness of SPM Module}
We first visualize learned SPM. 
We experiment with an extreme case to learn image translation between a set of squares and circles. The position, color and size are random. 
The results in Fig.~\ref{fig:toy} demonstrate that our SPM module produces reasonable shapes. Remaining visual artifacts are further reduced by the refinement module. 

\def\width0{0.20}
\begin{figure}[t]
	\begin{center}
		\begin{tabular}
			{@{\hspace{0.0mm}}c@{\hspace{2mm}}c@{\hspace{1mm}}c@{\hspace{1mm}}c@{\hspace{0.0mm}}}
			
			\fbox{\includegraphics[width=\width0\linewidth]{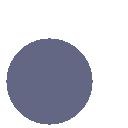}}&
			\fbox{\includegraphics[width=\width0\linewidth]{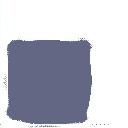}}&
			\fbox{\includegraphics[width=\width0\linewidth]{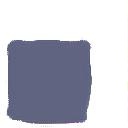}}&
			\fbox{\includegraphics[width=\width0\linewidth]{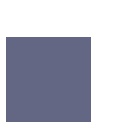}}\\
			
			\fbox{\includegraphics[width=\width0\linewidth]{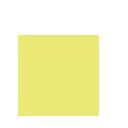}}&
			\fbox{\includegraphics[width=\width0\linewidth]{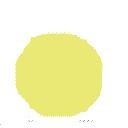}}&
			\fbox{\includegraphics[width=\width0\linewidth]{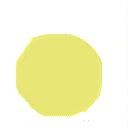}}&
			\fbox{\includegraphics[width=\width0\linewidth]{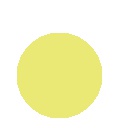}}\\
			Input & $SPM$ & $SPM+R$ & Ground Truth \\
		\end{tabular}
	\end{center}
	\vspace{-0.1in}
	\caption{Results on a synthetic dataset. $SPM$ indicates results generated by spontaneous motion module, and $SPM+R$ denotes final refinement results.}
	\vspace{-0.05in}
	\label{fig:toy}
\end{figure}

\vspace{-0.1in}	
\paragraph{Roles of Different Modules}
For the spontaneous motion module, we aim to generate reasonable geometric movement, e.g. lips stretched for smiling faces. For the refinement module, it further suppresses noise and adds more texture on deformation results to make images look more realistic.
A few intermediate and final `smiling' results produced from these modules under different resolutions ($128 \times 128$ to $256 \times 256$) are shown in Fig. \ref{fig:hires_result}. Effects from these two stages are clearly and respectively demonstrated. 

Besides, to further study the roles of different modules, we train our framework with no spontaneous motion module (\textbf{No\_M}) and no refinement module (\textbf{No\_R}) respectively to see how results are altered. We show results in Fig. \ref{fig:warp_refine} and the quantitative comparison in Tab.~\ref{tab:quan}. Without motion estimation, the geometric shape of images are wrong. The effect is like pasting patterns from the target domain to specific regions. Without the final refinement, results may contain distortions (right face in the 1st example) and artifacts (nose in the 2nd sample). Images also lack details to be a smiling face.

These experiments manifest the usefulness of both modules and our framework leverages their advantages.

\def\width1{0.24}
\begin{figure}[t]
	\begin{center}
		\begin{tabular}
			{@{\hspace{0.0mm}}c@{\hspace{2mm}}c@{\hspace{1mm}}c@{\hspace{1mm}}c@{\hspace{0.0mm}}}
			\includegraphics[width=\width1\linewidth]{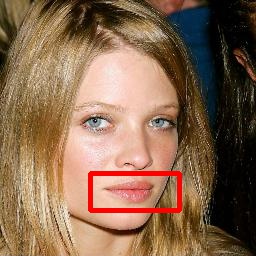}&
			\includegraphics[width=\width1\linewidth]{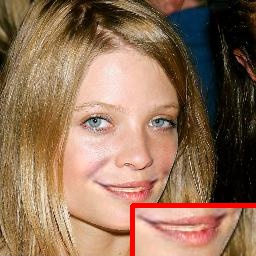}&
			\includegraphics[width=\width1\linewidth]{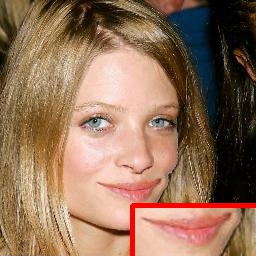}&
			\includegraphics[width=\width1\linewidth]{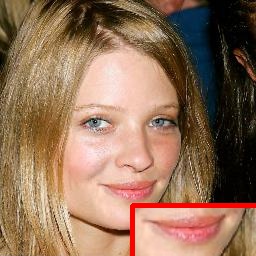}\\
			
			\includegraphics[width=\width1\linewidth]{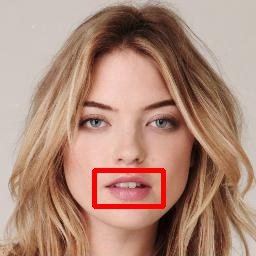}&
			\includegraphics[width=\width1\linewidth]{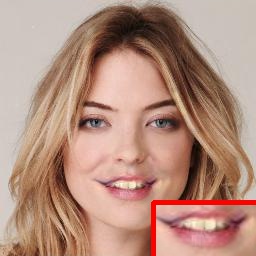}&
			\includegraphics[width=\width1\linewidth]{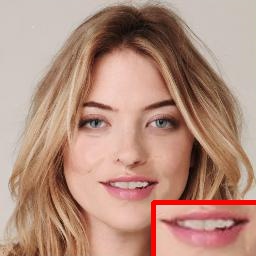}&
			\includegraphics[width=\width1\linewidth]{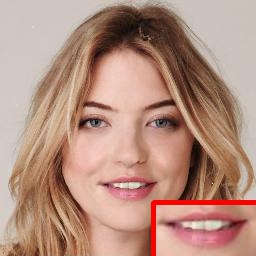}\\
			
			Input & No\_M & No\_R & Full \\
			
		\end{tabular}
	\end{center}
	\vspace{-0.1in}
	\caption{Ablation study on important modules. \textbf{No\_M} indicates no spontaneous motion estimation, \textbf{No\_R} refers to no refinement. \textbf{Full} indicates our final full framework.\vspace{-0.1in}}
	\label{fig:warp_refine}	
\end{figure}

\vspace{-0.1in}		
\paragraph{SPM Field in Different Conditions}
Motion patterns for the same face expression are generally similar even with different input images. For example, non-smiling to smiling faces need to `stretch' pixels of lips. Taking the RaFD dataset as an example, motion fields for different emotions are visualized in Fig. \ref{fig:warp_field}. They tell different parts of faces required to be updated to achieve ideal facial expression.

\def\width2{0.12}
\begin{figure*}[t]
	\begin{center}
		\begin{tabular}
			{@{\hspace{0.0mm}}c@{\hspace{2mm}}c@{\hspace{1mm}}c@{\hspace{1mm}}c@{\hspace{1mm}}c@{\hspace{1mm}}c@{\hspace{1mm}}c@{\hspace{1mm}}c@{\hspace{0.0mm}}}
			\includegraphics[width=\width2\linewidth]{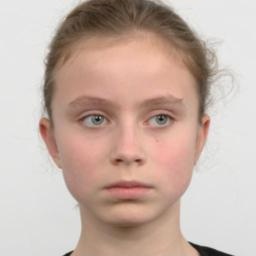}&
			\includegraphics[width=\width2\linewidth]{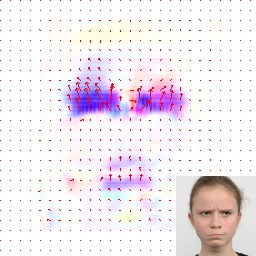}&
			\includegraphics[width=\width2\linewidth]{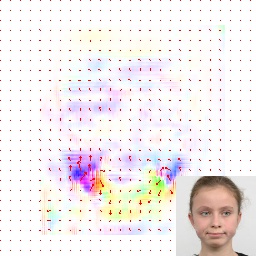}&
			\includegraphics[width=\width2\linewidth]{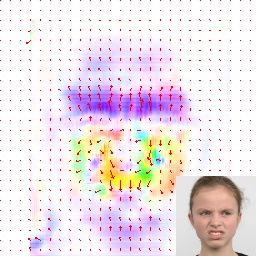}&
			\includegraphics[width=\width2\linewidth]{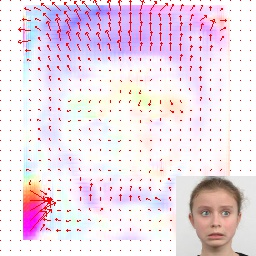}&
			\includegraphics[width=\width2\linewidth]{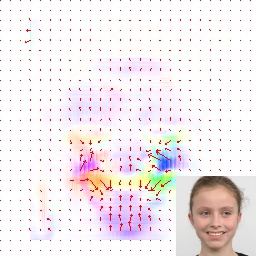}&
			\includegraphics[width=\width2\linewidth]{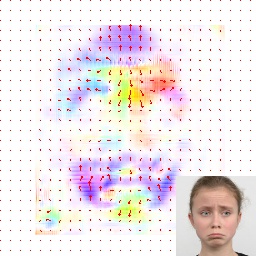}&
			\includegraphics[width=\width2\linewidth]{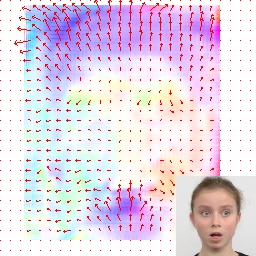}\\
			
			\includegraphics[width=\width2\linewidth]{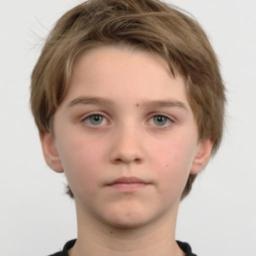}&
			\includegraphics[width=\width2\linewidth]{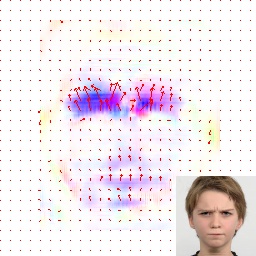}&
			\includegraphics[width=\width2\linewidth]{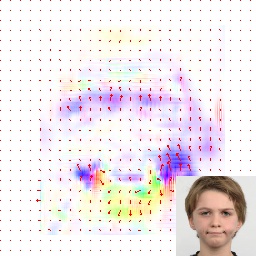}&
			\includegraphics[width=\width2\linewidth]{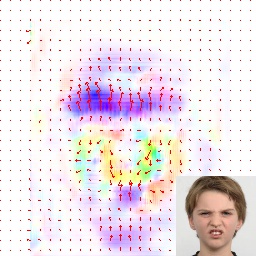}&
			\includegraphics[width=\width2\linewidth]{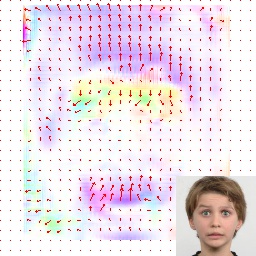}&
			\includegraphics[width=\width2\linewidth]{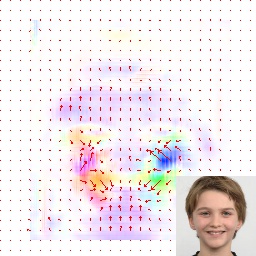}&
			\includegraphics[width=\width2\linewidth]{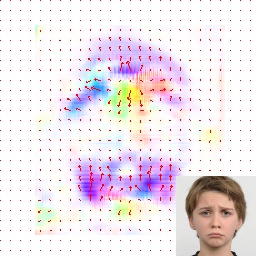}&
			\includegraphics[width=\width2\linewidth]{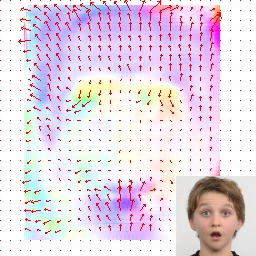}\\
			
			\includegraphics[width=\width2\linewidth]{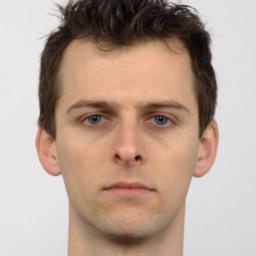}&
			\includegraphics[width=\width2\linewidth]{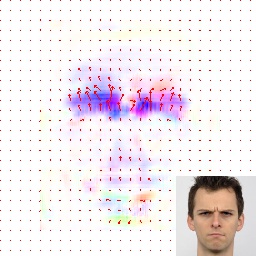}&
			\includegraphics[width=\width2\linewidth]{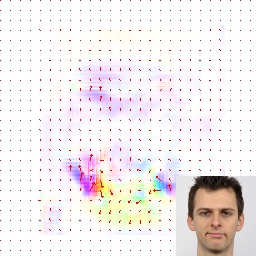}&
			\includegraphics[width=\width2\linewidth]{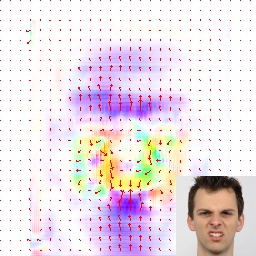}&
			\includegraphics[width=\width2\linewidth]{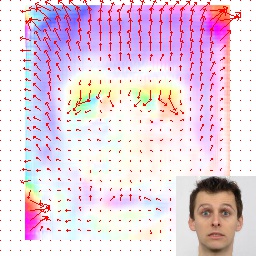}&
			\includegraphics[width=\width2\linewidth]{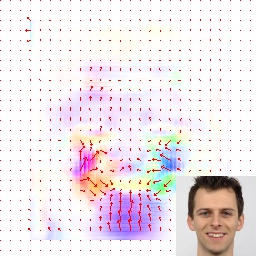}&
			\includegraphics[width=\width2\linewidth]{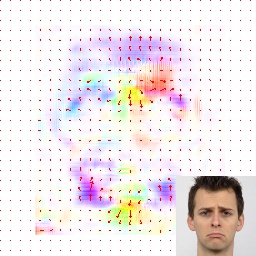}&
			\includegraphics[width=\width2\linewidth]{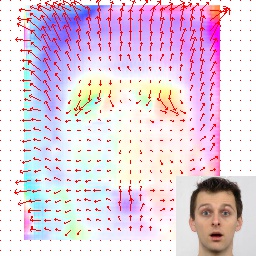}\\
			Input & Angry & Contemptuous & Disgusted & Fearful & Happy & Sad & Surprised \\
		\end{tabular}
	\end{center}
	\caption{Spontaneous motion field visualization under different conditions, where ``Input" denotes neutral faces (best viewed in color).}
	\label{fig:warp_field}
\end{figure*}

\def\widthh{0.155}
\begin{figure*}[t]
	\begin{center}
		\begin{tabular}
			{@{\hspace{0.0mm}}c@{\hspace{2mm}}c@{\hspace{1mm}}c@{\hspace{1mm}}c@{\hspace{1mm}}c@{\hspace{1mm}}c@{\hspace{0.0mm}}}
			\includegraphics[width=\widthh\linewidth]{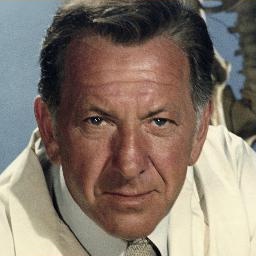}&
			\includegraphics[width=\widthh\linewidth]{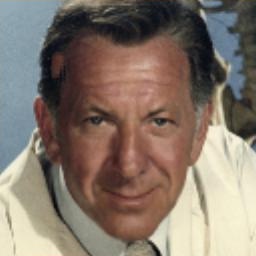}&
			\includegraphics[width=\widthh\linewidth]{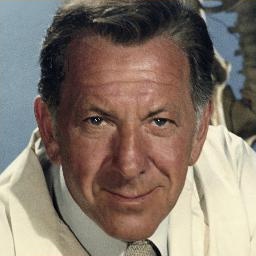}&
			\includegraphics[width=\widthh\linewidth]{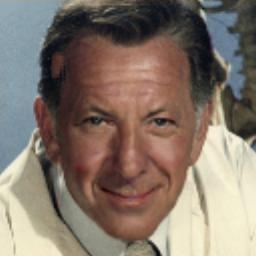}&
			\includegraphics[width=\widthh\linewidth]{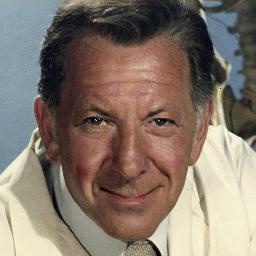}&
			\includegraphics[width=\widthh\linewidth]{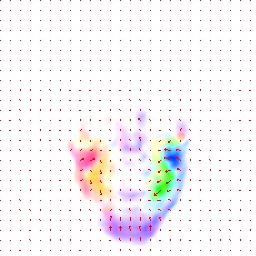}\\
			
			\includegraphics[width=\widthh\linewidth]{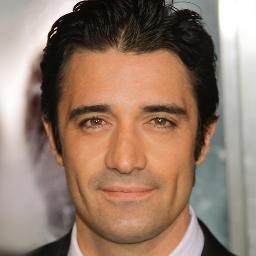}&
			\includegraphics[width=\widthh\linewidth]{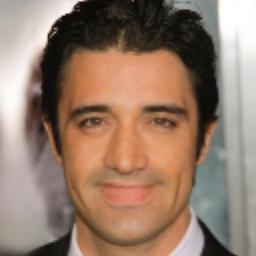}&
			\includegraphics[width=\widthh\linewidth]{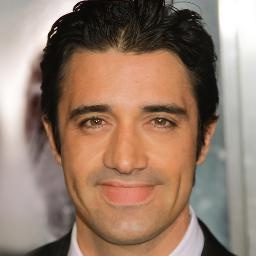}&
			\includegraphics[width=\widthh\linewidth]{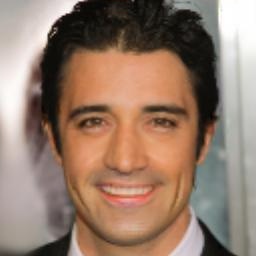}&
			\includegraphics[width=\widthh\linewidth]{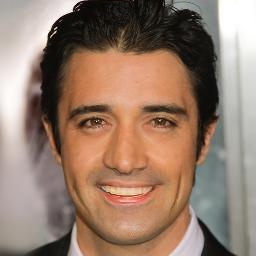}&
			\includegraphics[width=\widthh\linewidth]{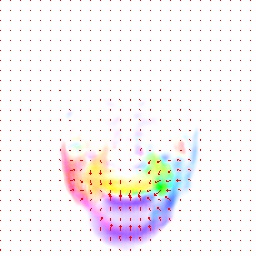}\\
			
			Input & $SPM_l$ & $SPM_h$ & $R_l$ & $R_h$ & $w$ \\
			
		\end{tabular}
		
	\end{center}
	\caption{Intermediate and final results output from the higher resolution framework, where $SPM_l$ and $SPM_h$ refer to LR and HR deformation results. $R_l$ and $R_h$ are LR and HR refinement results. $w$ indicates corresponding high resolution motion fields.	\vspace{-0.1in}}
	\label{fig:hires_result}
\end{figure*}

\def\widthb{0.14}
\begin{figure*}
	\begin{center}
		\begin{tabular}
			{@{\hspace{0.0mm}}c@{\hspace{2mm}}c@{\hspace{1mm}}c@{\hspace{1mm}}c@{\hspace{1mm}}c@{\hspace{1mm}}c@{\hspace{1mm}}c@{\hspace{0.0mm}}}
			&
			\includegraphics[width=\widthb\linewidth]{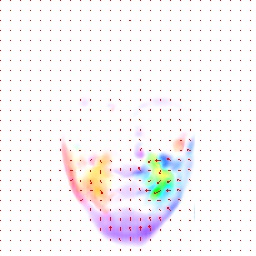}&
			\includegraphics[width=\widthb\linewidth]{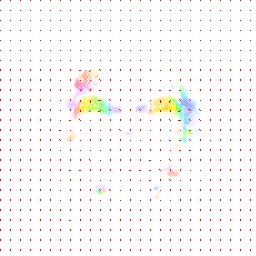}&
			\includegraphics[width=\widthb\linewidth]{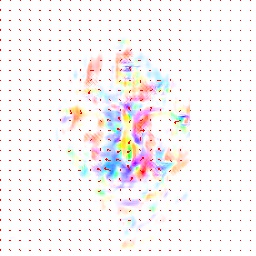}&
			\includegraphics[width=\widthb\linewidth]{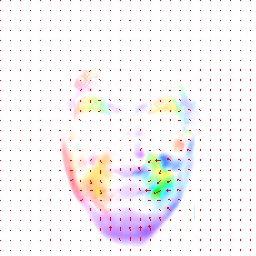} &
			\includegraphics[width=\widthb\linewidth]{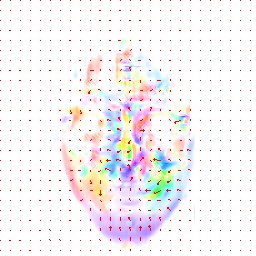} &
			\includegraphics[width=\widthb\linewidth]{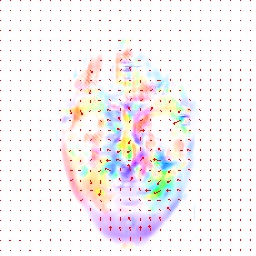} \\
			
			\includegraphics[width=\widthb\linewidth]{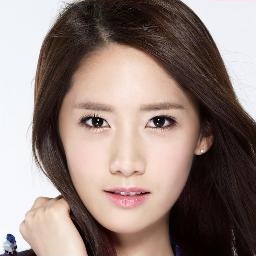}&
			\includegraphics[width=\widthb\linewidth]{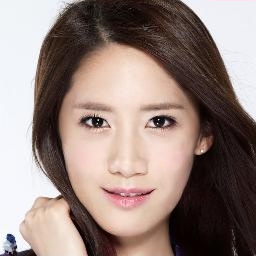}&
			\includegraphics[width=\widthb\linewidth]{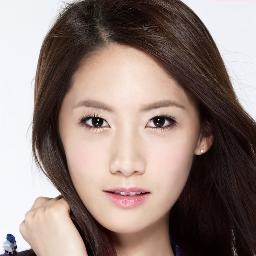}&
			\includegraphics[width=\widthb\linewidth]{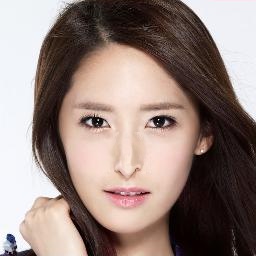}&
			\includegraphics[width=\widthb\linewidth]{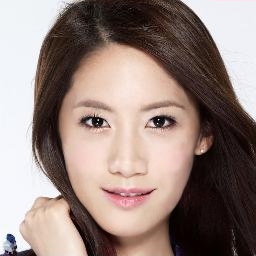} &
			\includegraphics[width=\widthb\linewidth]{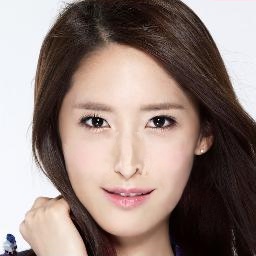} &
			\includegraphics[width=\widthb\linewidth]{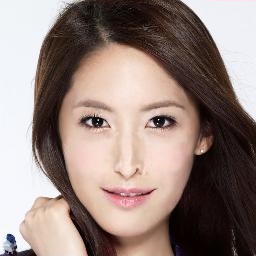} \\
			
			Input & $M_s$ & $M_e$ & $M_n$ & $M_s+M_e$ & $M_s+M_n$ & $M_s+M_e+M_n$ \\
			
		\end{tabular}
		
	\end{center}
	\vspace{-0.12in}
	\caption{Motion field basis combination. First row: motion fields under different conditions. Second row: deformation results by applying corresponding motion fields. $M_s$: `smiling' transform, $M_e$: `arched\_eyebrow' transform, $M_n$: `pointy\_nose' transform. $M_s+M_e$, $M_s+M_n$, $M_s+M_e+M_n$ are with two or three corresponding motion field combination. }
	\label{fig:flow_comb}
\end{figure*}

\def\widtho{0.155}
\begin{figure*}
	\begin{center}
		\begin{tabular}
			{@{\hspace{0.0mm}}c@{\hspace{1mm}}c@{\hspace{2mm}}c@{\hspace{1mm}}c@{\hspace{1mm}}c@{\hspace{1mm}}c@{\hspace{1mm}}c@{\hspace{0.0mm}}}
			\rotatebox{90}{\hspace{6.0mm} Smiling} &
			\includegraphics[width=\widtho\linewidth]{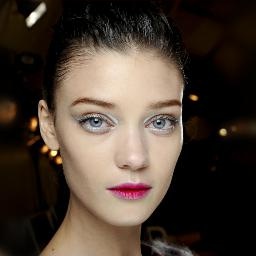}&
			\includegraphics[width=\widtho\linewidth]{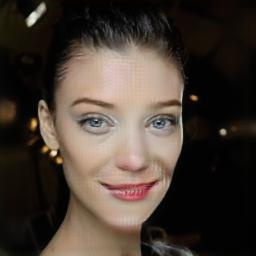}&
			\includegraphics[width=\widtho\linewidth]{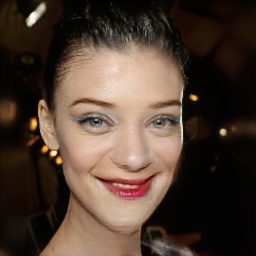}&
			\includegraphics[width=\widtho\linewidth]{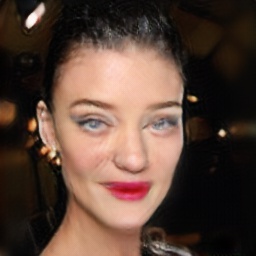}&
			\includegraphics[width=\widtho\linewidth]{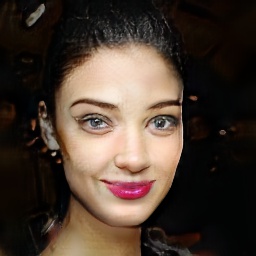}&
			\includegraphics[width=\widtho\linewidth]{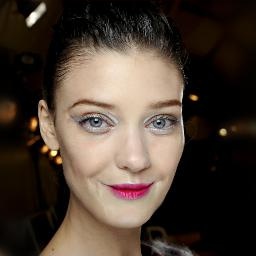}\\
			
			\rotatebox{90}{\hspace{5.5mm} Big\_Nose} &
			\includegraphics[width=\widtho\linewidth]{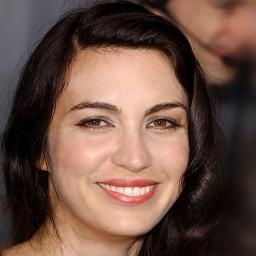}&
			\includegraphics[width=\widtho\linewidth]{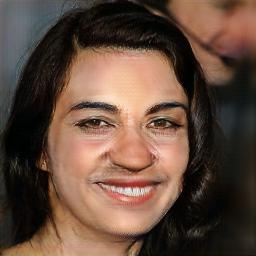}&
			\includegraphics[width=\widtho\linewidth]{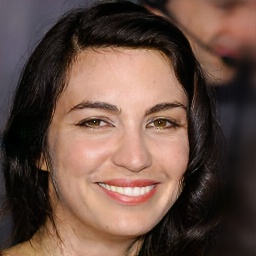}&
			\includegraphics[width=\widtho\linewidth]{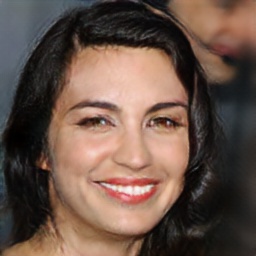}&
			\includegraphics[width=\widtho\linewidth]{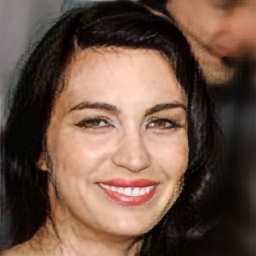}&
			\includegraphics[width=\widtho\linewidth]{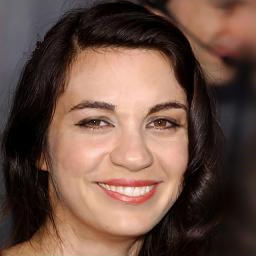}\\
			
			\rotatebox{90}{\hspace{3.7mm} Pointy\_Nose} &
			\includegraphics[width=\widtho\linewidth]{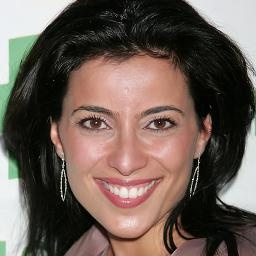}&
			\includegraphics[width=\widtho\linewidth]{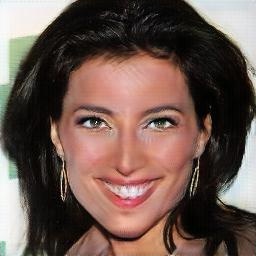}&
			\includegraphics[width=\widtho\linewidth]{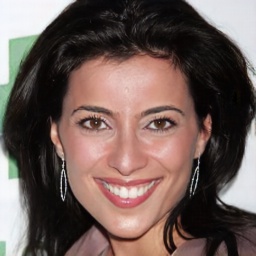}&
			\includegraphics[width=\widtho\linewidth]{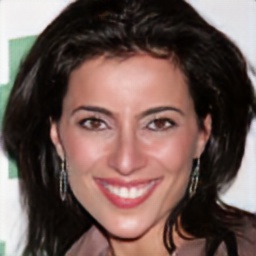}&
			\includegraphics[width=\widtho\linewidth]{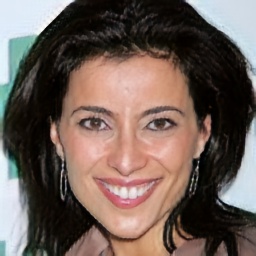}&
			\includegraphics[width=\widtho\linewidth]{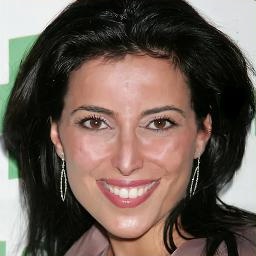}\\
			
			\rotatebox{90}{\hspace{0.5mm} Arched\_Eyebrow} &
			\includegraphics[width=\widtho\linewidth]{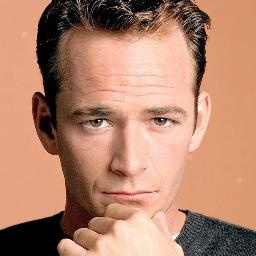}&
			\includegraphics[width=\widtho\linewidth]{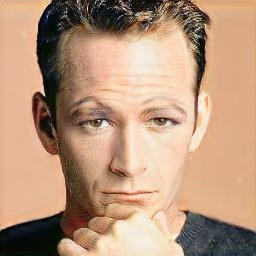}&
			\includegraphics[width=\widtho\linewidth]{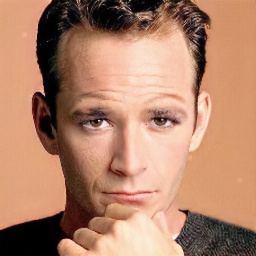}&
			\includegraphics[width=\widtho\linewidth]{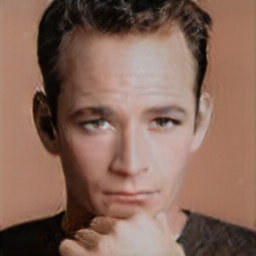}&
			\includegraphics[width=\widtho\linewidth]{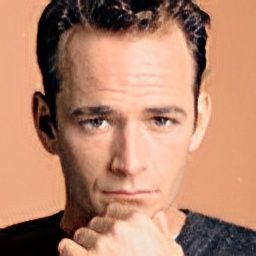}&
			\includegraphics[width=\widtho\linewidth]{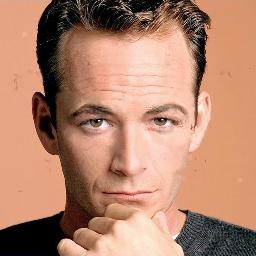}\\

			& Input & StarGAN~\cite{choi2018stargan} & CycleGAN~\cite{zhu2017unpaired} & Ganimorph~\cite{gokaslan2018improving} & MUNIT~\cite{huang2018multimodal}  & Ours \\  
			
		\end{tabular}
	\end{center}
	\vspace{-0.1in}
	\caption{Visual quality comparison on the CelebA dataset.	\vspace{-0.1in}}
	\label{fig:single_t}
\end{figure*}

\vspace{-0.1in}
\paragraph{Spontaneous Motion Field Basis Combination}
Transformation varies when applying different motions to the same image. Since motion fields are independent, we can combine motion fields with simple addition operations. By adding differently learned motion fields, we achieve rough expression combination without re-training the network, under the condition that the combined transformation does not conflict with each other. We demonstrate the effect of combination in Fig. \ref{fig:flow_comb}. The difference in generated micro-expression is very useful for fine face attribute creation.

\subsection{Comparisons}
We compare with several prevalent methods in image translation. They are StarGAN~\cite{choi2018stargan}, CycleGAN~\cite{zhu2017unpaired}, MUNIT~\cite{huang2018multimodal} and Ganimorph~\cite{gokaslan2018improving}. StarGAN \cite{choi2018stargan} is the first framework for multi-condition image translation. CycleGAN~\cite{zhu2017unpaired} and MUNIT~\cite{huang2018multimodal} are important methods in image translation. Ganimorph~\cite{gokaslan2018improving} is a geometry-aware framework based on CycleGAN~\cite{zhu2017unpaired}, which is another solution to geometric transformation across domains in image translation.

\vspace{-0.05in}
\subsubsection{Visual Comparison}
\vspace{-0.05in}
We conduct experiments on the two datasets for comparative evaluation. On the CelebA dataset, we treat each attribute $x$ transformation as a two-domain translation from non-$x$ to $x$. Fig.~\ref{fig:single_t} shows that CycleGAN, MUNIT, and Ganimorph cannot capture domain information when the attribute transformation is subtle, like `Big\_Nose' and `Pointy\_Nose'. They tend to reconstruct the input image instead. Both StarGAN and our method handle such subtle domain translation thanks to the domain classifiers. 

Our method better tackles geometry variation and image misalignment. For other attributes like `smiling', though all previous methods transform source images to target domain, various types of geometric deformation lead to quality difference on results, causing noticeable ghosting or artifacts. Our method, contrarily, alleviates this issue.

For the RaFD dataset (Fig.~\ref{fig:mul_t}), similarly, StarGAN handles domain transformation and yet are with room to improve details and geometric shapes, especially for the `happy' expression. Our framework satisfies target conditions better thanks to our explicit spontaneous motion module and our two domain classifiers for training.

\def\widthm{0.12}
\begin{figure*}
	\begin{center}
		\begin{tabular}
			{@{\hspace{0.0mm}}c@{\hspace{1.5mm}}c@{\hspace{1.5mm}}c@{\hspace{1.0mm}}c@{\hspace{1.0mm}}c@{\hspace{1.0mm}}c@{\hspace{1.0mm}}c@{\hspace{1.0mm}}c@{\hspace{1.0mm}}c@{\hspace{0mm}}}
			\rotatebox{90}{\hspace{4.0mm} \footnotesize StarGAN} &
			\includegraphics[width=\widthm\linewidth]{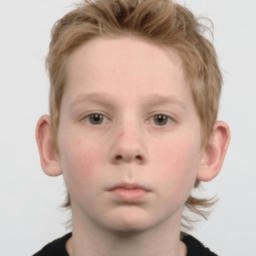} &
			\includegraphics[width=\widthm\linewidth]{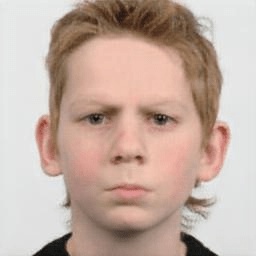}&
			\includegraphics[width=\widthm\linewidth]{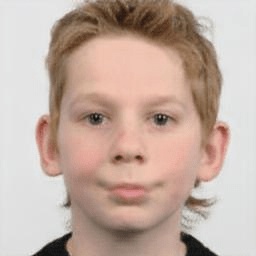}&
			\includegraphics[width=\widthm\linewidth]{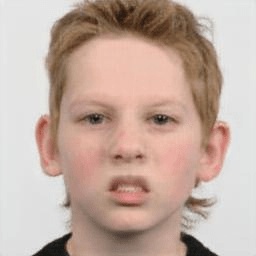}&
			\includegraphics[width=\widthm\linewidth]{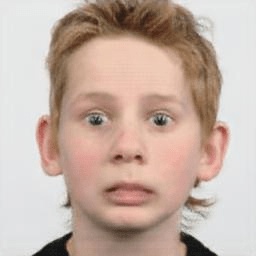}&
			\includegraphics[width=\widthm\linewidth]{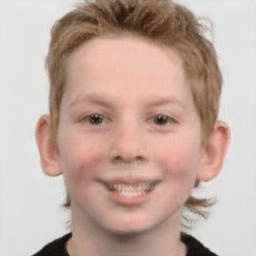}&
			\includegraphics[width=\widthm\linewidth]{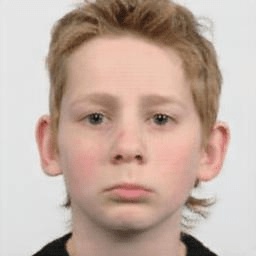}&
			\includegraphics[width=\widthm\linewidth]{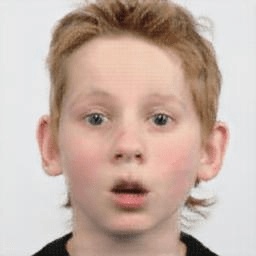}\\
			
			\rotatebox{90}{\hspace{4.0mm} \footnotesize CycleGAN} &
			\includegraphics[width=\widthm\linewidth]{fig_compare/multi_attr/inp_70_4.jpg} &
			\includegraphics[width=\widthm\linewidth]{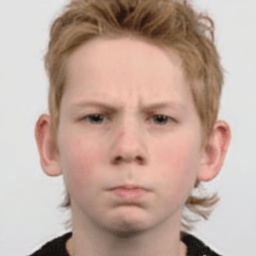}&
			\includegraphics[width=\widthm\linewidth]{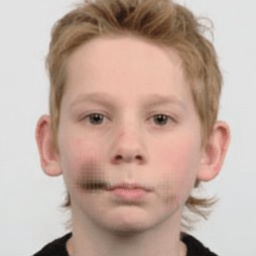}&
			\includegraphics[width=\widthm\linewidth]{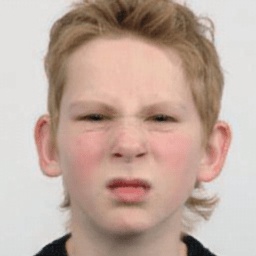}&
			\includegraphics[width=\widthm\linewidth]{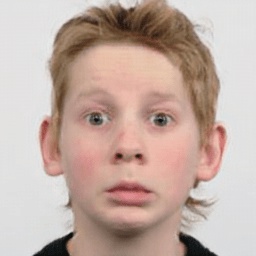}&
			\includegraphics[width=\widthm\linewidth]{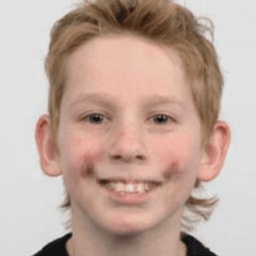}&
			\includegraphics[width=\widthm\linewidth]{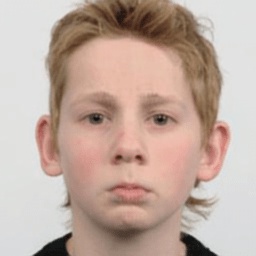}&
			\includegraphics[width=\widthm\linewidth]{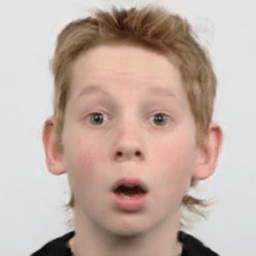}\\
			
			\rotatebox{90}{\hspace{4.0mm} \footnotesize Ganimorph} &
			\includegraphics[width=\widthm\linewidth]{fig_compare/multi_attr/inp_70_4.jpg} &
			\includegraphics[width=\widthm\linewidth]{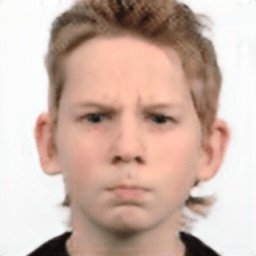}&
			\includegraphics[width=\widthm\linewidth]{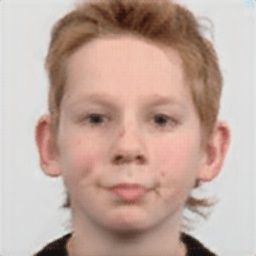}&
			\includegraphics[width=\widthm\linewidth]{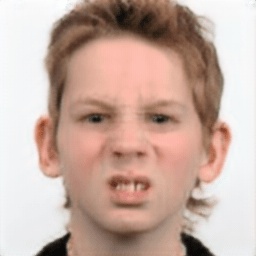}&
			\includegraphics[width=\widthm\linewidth]{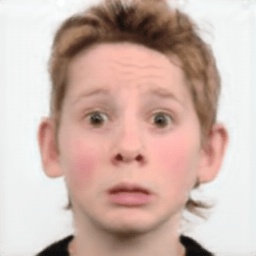}&
			\includegraphics[width=\widthm\linewidth]{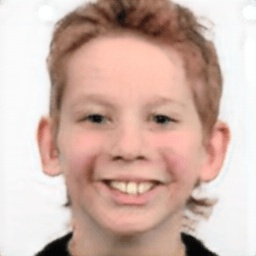}&
			\includegraphics[width=\widthm\linewidth]{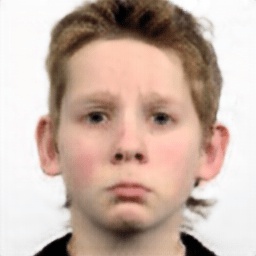}&
			\includegraphics[width=\widthm\linewidth]{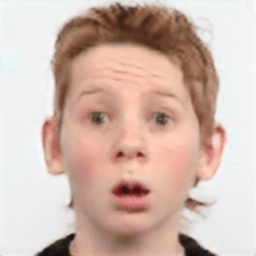}\\
			
			\rotatebox{90}{\hspace{6.0mm} \footnotesize Ours} &
			\includegraphics[width=\widthm\linewidth]{fig_compare/multi_attr/inp_70_4.jpg} &
			\includegraphics[width=\widthm\linewidth]{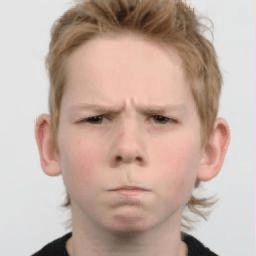}&
			\includegraphics[width=\widthm\linewidth]{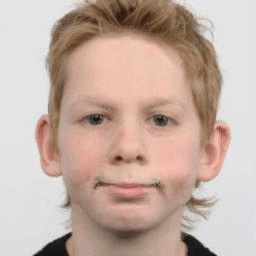}&
			\includegraphics[width=\widthm\linewidth]{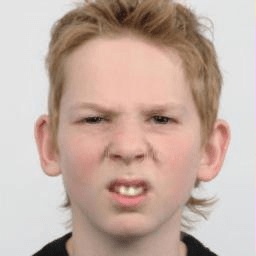}&
			\includegraphics[width=\widthm\linewidth]{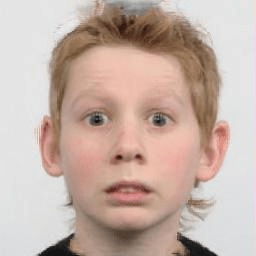}&
			\includegraphics[width=\widthm\linewidth]{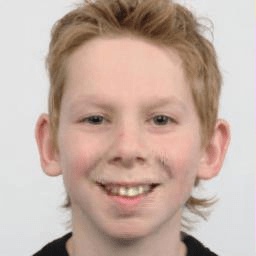}&
			\includegraphics[width=\widthm\linewidth]{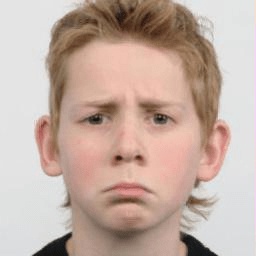}&
			\includegraphics[width=\widthm\linewidth]{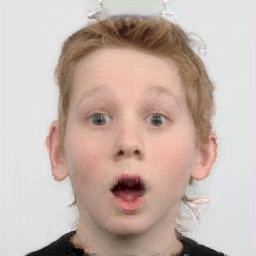}\\
			& Input & Angry & Contemptuous & Disgusted & Fearful & Happy & Sad & Surprised \\
		\end{tabular}
	\end{center}
	\vspace{-0.1in}
	\caption{Visual quality comparison on different expressions on the RaFD dataset.}
	\label{fig:mul_t}
\end{figure*}

\vspace{-0.05in}
\subsubsection{Quantitative Comparison}
\vspace{-0.05in}
\paragraph{Distribution Discrepancy}
To evaluate generated faces quantitatively, we extract features with a deep face feature extractor VGGFace2~\cite{cao2018vggface2} and use FID~\cite{heusel2017gans} to measure feature distribution discrepancy between real and generated faces.
For each attribute, we first extract feature ${F_r}$ from real faces with such an attribute in test set, and then extract features ${F_g}_{i}$ from translated images (to this attribute) by each method to be compared. 

We calculate FID between ${F_r}$ and ${F_g}_{i}$ for each method. Results in Tab.~\ref{tab:quan} demonstrate that our framework achieves the lowest FID score among all methods, which indicates that the feature distribution of our generated images is closest to that of real images.

\vspace{-0.1in}
\begin{table}[h]
	\begin{center}
		{\tabcolsep0.05in
			\small{
				\begin{tabular}
					{c|cccc|c}
					\hline
					Methods / ($\times1e3$) &  S $\downarrow$ & BN $\downarrow$ & PN $\downarrow$ & AE $\downarrow$ & Acc.(\%) $\uparrow$\\
					\hline
					StarGAN  & ~3.676 & 7.875  & 6.933 & 3.751 & 95.67\\
					CycleGAN & ~4.011 & 5.262 & 4.886 & 4.171 & 91.23 \\
					Ganimorph & ~4.689 & 5.645 & 5.129 & 5.570 & 86.25 \\
					MUNIT & ~5.189 & 5.551 & 4.761 & 5.271 & 81.43\\
					\hline
					Ours (No\_M) & ~3.224 & 6.051 & 5.682 & 4.667 & 88.29 \\
					Ours (No\_R) & ~3.022 & 6.505 & 5.894 & 3.911 & 90.96 \\
					Ours (Full) & ~\textbf{2.907} & \textbf{5.137} & \textbf{4.704} & \textbf{3.678} & \textbf{97.85} \\
					\hline
					Real & - & - & - & - & 98.75 \\
					\hline
			\end{tabular}}
		}
	\end{center}
	\vspace{-0.05in}
	\caption{Quantitative comparison in terms of distribution discrepancy and classification accuracy. For each facial attribute, we compare FID scores among methods. ``S'', ``BN'', ``PN'' and ``AE'' indicate \emph{Smiling}, \emph{Big\_Nose}, \emph{Pointy\_Nose} and \emph{Arched\_Eyebrow} respectively, while ``Acc." refers to classification accuracy.	\vspace{-0.01in}}
	\label{tab:quan}

\end{table}

\vspace{-0.05in}
\paragraph{Classification Accuracy}
Following \cite{choi2018stargan}, we compute the classification accuracy of facial expression on generated images. We first train a facial expression classifier with ResNet-18~\cite{He_2016_CVPR} on the RaFD dataset with the train set. We achieve near-perfect accuracy of 98.75\% on test set.
Then we apply this well-trained classifier to compute classification accuracy on synthesized images output from different methods. The results in Table~\ref{tab:quan} indicate that we achieve the best results in terms of classification accuracy. StarGAN works also very well benefited by its domain classification framework.

\subsubsection{User Study}
We also conduct user study for method comparison among 101 subjects, with 21 groups of generated samples. 
Given an input image, subjects are instructed to choose the best item based on quality of attribute transfer, perceptual realism, and preservation of identity. 
The results in Table~\ref{tab:user_study} demonstrate that our method performs best among different facial attribute transformation methods, while StarGAN~\cite{choi2018stargan} performs well for subtle facial attribute (\eg \emph{Big\_Nose}) transformation and CycleGAN~\cite{zhu2017unpaired} yields decent output on obvious attributes (\eg \emph{Arched\_Eyebrow}, \emph{Smiling}).

\begin{table}[t]
	\begin{center}
		\small{
			\begin{tabular}
				{c|cccc}
				\hline
				Methods &  S (\%) & BN (\%) & PN (\%) & AE (\%) \\
				\hline
				StarGAN & 10.17 & 25.66 & 16.83 & 24.24 \\
				CycleGAN & 14.41 & 8.85 & 14.85 & 28.79 \\
				Ganimorph & 5.93 & 5.31 & 6.93 & 1.52 \\
				MUNIT & 11.44 & 7.96 & 11.88 & 3.03 \\
				Ours & \textbf{58.05} & \textbf{52.21} & \textbf{49.50} & \textbf{42.42} \\
				\hline
		\end{tabular}}
	\end{center}
	\vspace{-0.08in}
	\caption{User study for different attribute translation among methods. The value refers to the ratio of selecting as best item.\vspace{-0.01in}}
	\label{tab:user_study}
\end{table}

\section{Conclusion}
In this paper, we have introduced geometric deformation into image translation frameworks. We proposed spontaneous motion estimation module followed by refinement to fix remaining artifacts in deformation results. Extensive experiments manifest the effectiveness of our proposed framework. It achieves promising results for image translation and enables new visualization and applications. Our method may also shed lights on geometric-aware image translation.

\clearpage
{\small
	\bibliographystyle{ieee}
	\bibliography{adspm}
}
\clearpage
\section{Appendix}
\vspace{-0.051in}
\subsection{More Discussions}
\vspace{-0.05in}
\subsubsection{Attention Mask Learning}
\vspace{-0.05in}
We learn attention masks to filter out unnecessary noises and focus on essential regions during transformation. 
Taking the attribute (Smiling, Arched\_eyebrow, Big\_Nose) transforms in CelebA dataset as examples, the generated attention mask is shown in Fig.~\ref{fig:att_mask}. 
We can see that for `Smiling' transformation, the attention area is mouth, cheekbones and related region. `Arched\_eyebrow' transformation will focus on eyebrow region and `Big\_Nose' will focus on two sides of nose. 
\def\widthc{0.22}
\begin{figure}[h]
	\begin{center}
		\begin{tabular}
			{@{\hspace{0.0mm}}c@{\hspace{1.5mm}}c@{\hspace{1.0mm}}c@{\hspace{1.0mm}}c@{\hspace{1.0mm}}c@{\hspace{1.0mm}}c@{\hspace{0mm}}}
			\rotatebox{90}{\hspace{3.0mm} Smiling} &
			\includegraphics[width=\widthc\linewidth]{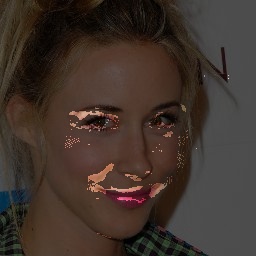}&
			\includegraphics[width=\widthc\linewidth]{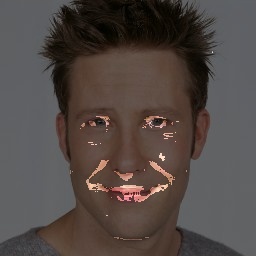}&
			\includegraphics[width=\widthc\linewidth]{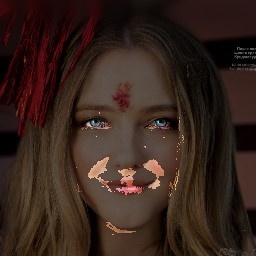}&
			\includegraphics[width=\widthc\linewidth]{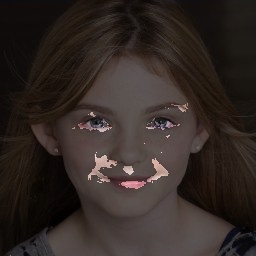}\\
			
			\rotatebox{90}{\hspace{2.0mm} Eyebrow} &
			\includegraphics[width=\widthc\linewidth]{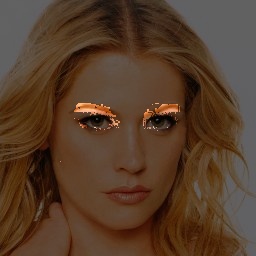}&
			\includegraphics[width=\widthc\linewidth]{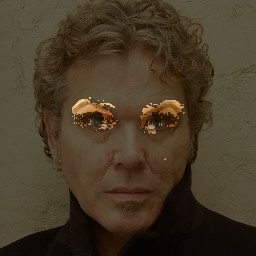}&
			\includegraphics[width=\widthc\linewidth]{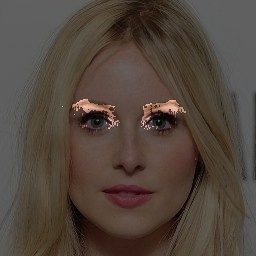}&
			\includegraphics[width=\widthc\linewidth]{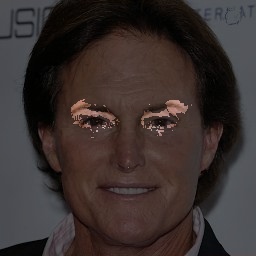}\\
			
			\rotatebox{90}{\hspace{4.5mm} Nose} &
			\includegraphics[width=\widthc\linewidth]{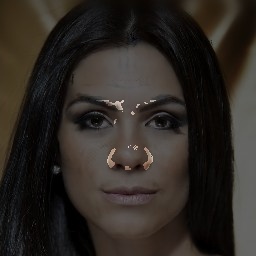}&
			\includegraphics[width=\widthc\linewidth]{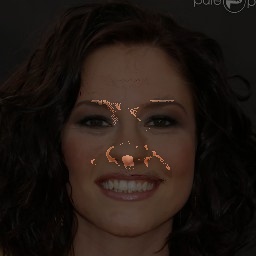}&
			\includegraphics[width=\widthc\linewidth]{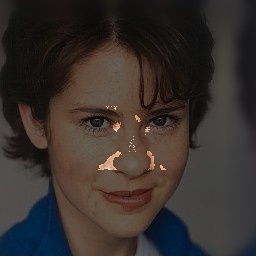}&
			\includegraphics[width=\widthc\linewidth]{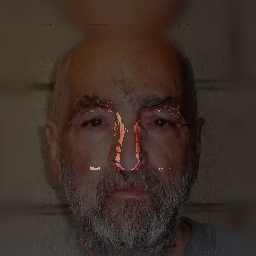}\\
			
		\end{tabular}
		
	\end{center}\vspace{-0.1in}
	\caption{The generated Attention mask in different transformations. The highlight area refers to the attention region, here we take `smiling', `Arched\_eyebrow' and `Big\_Nose' transformation as examples.}
	\label{fig:att_mask}\vspace{-0.2in}
\end{figure}

\subsubsection{Other Application}
\vspace{-0.05in}
\paragraph{Face to Bitmoji}
We also conduct experiments on other image translation applications to test the effectiveness of our framework. Specifically, we evaluate our framework in real face to bitmoji face application, since the two domains are quite different in both texture and geometric shape. The results are demonstrated in Fig.~\ref{fig:cartoon_bitmoji}. 
From the results we can see that spontaneous motion module will firstly deform input faces to bitmoji faces in geometric shape ($\textbf{SPM}$), then the refinement module render the deformed results with bitmoji textures to get the final results ($\textbf{SPM + R}$). This working pipeline may effectively get rid of distortions and artifacts caused by current generators with all aligned operators (\ie convolution / de-convolution layers).

\def\widthj{0.22}
\begin{figure}[h]
	\begin{center}
		\begin{tabular}
			{@{\hspace{0.0mm}}c@{\hspace{2mm}}c@{\hspace{1mm}}c@{\hspace{1mm}}c@{\hspace{0.0mm}}}

			\includegraphics[width=\widthj\linewidth]{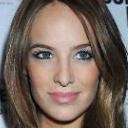}&
			\includegraphics[width=\widthj\linewidth]{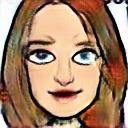}&
			\includegraphics[width=\widthj\linewidth]{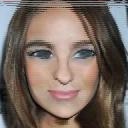}&
			\includegraphics[width=\widthj\linewidth]{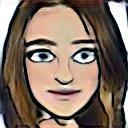}\\

			\includegraphics[width=\widthj\linewidth]{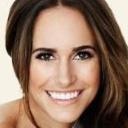}&
			\includegraphics[width=\widthj\linewidth]{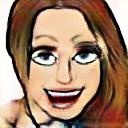}&
			\includegraphics[width=\widthj\linewidth]{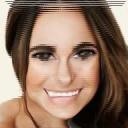}&
			\includegraphics[width=\widthj\linewidth]{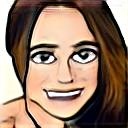}\\
			
			\includegraphics[width=\widthj\linewidth]{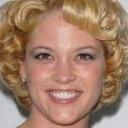}&
			\includegraphics[width=\widthj\linewidth]{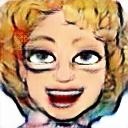}&
			\includegraphics[width=\widthj\linewidth]{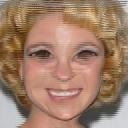}&
			\includegraphics[width=\widthj\linewidth]{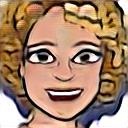}\\

			\includegraphics[width=\widthj\linewidth]{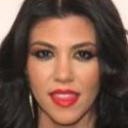}&
			\includegraphics[width=\widthj\linewidth]{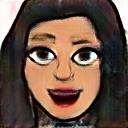}&
			\includegraphics[width=\widthj\linewidth]{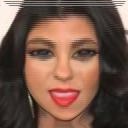}&
			\includegraphics[width=\widthj\linewidth]{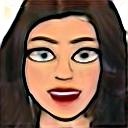}\\

			\includegraphics[width=\widthj\linewidth]{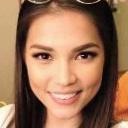}&
			\includegraphics[width=\widthj\linewidth]{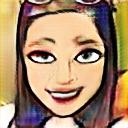}&
			\includegraphics[width=\widthj\linewidth]{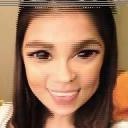}&
			\includegraphics[width=\widthj\linewidth]{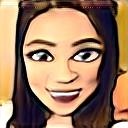}\\
			
			Input & StarGAN & $SPM$ & $SPM + R$ \\
			
		\end{tabular}
		
	\end{center}\vspace{-0.1in}
	\caption{Real faces to bitmoji faces transformation, where $SPM$ indicates deformed result and $SPM + R$ indicates refined deformed result.}
	\label{fig:cartoon_bitmoji}
\end{figure}

\subsubsection{High Resolution Results}
\vspace{-0.05in}
We demonstrate higher resolution, \ie $512 \times 512$ results in Fig. \ref{fig:hires512}. 
All results with different resolutions in our paper start from generating $128 \times 128$ images.
To generate $256 \times 256$ results, we adopt the training strategy on Sec 3.4 in our paper for one time, while we would up-sample twice with the same strategy for $512 \times 512$ image generation. 

\def\widthj{0.244}
\begin{figure*}[h]
	\begin{center}
		\begin{tabular}
			{@{\hspace{0.0mm}}c@{\hspace{0.5mm}}c@{\hspace{2mm}}c@{\hspace{0.5mm}}c@{\hspace{0.0mm}}}
			
			\includegraphics[width=\widthj\linewidth]{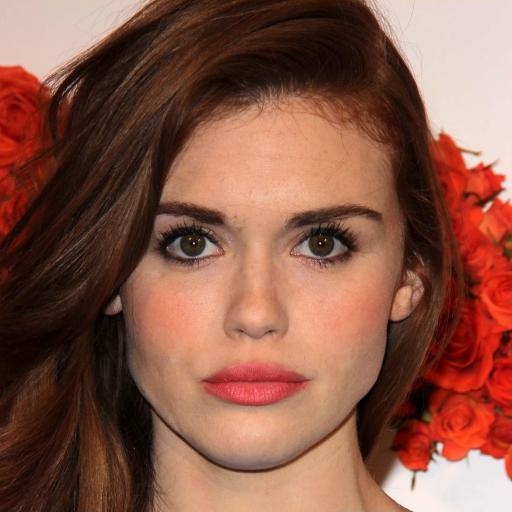}&
			\includegraphics[width=\widthj\linewidth]{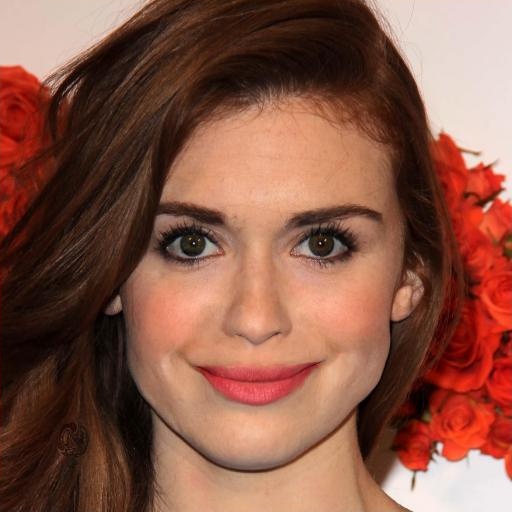}&
			
			\includegraphics[width=\widthj\linewidth]{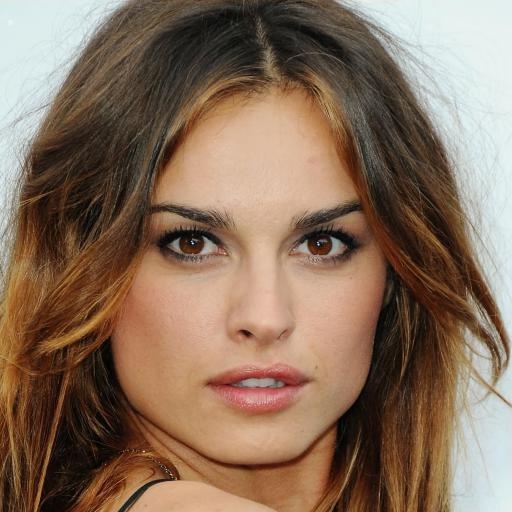}&
			\includegraphics[width=\widthj\linewidth]{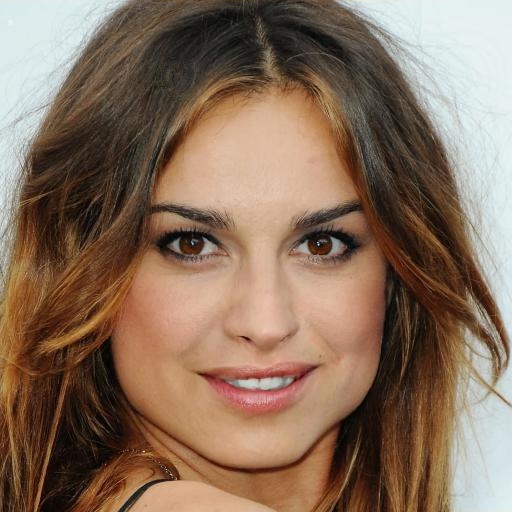}\\
			
			\includegraphics[width=\widthj\linewidth]{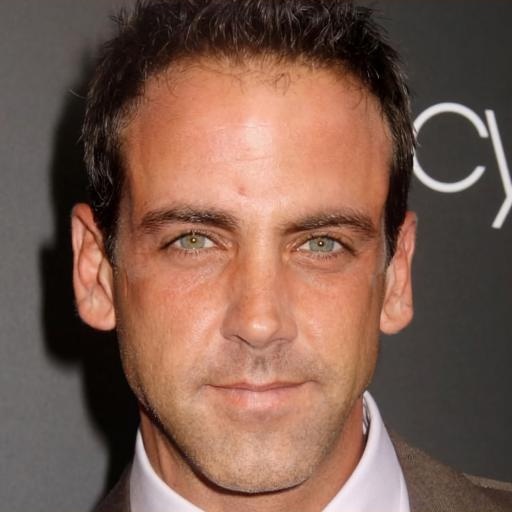}&
			\includegraphics[width=\widthj\linewidth]{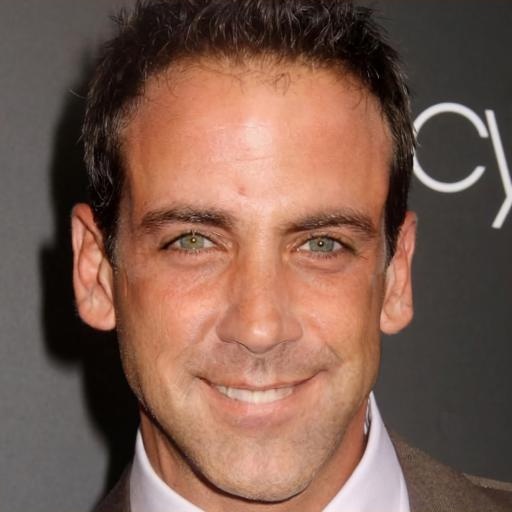}&
			
			\includegraphics[width=\widthj\linewidth]{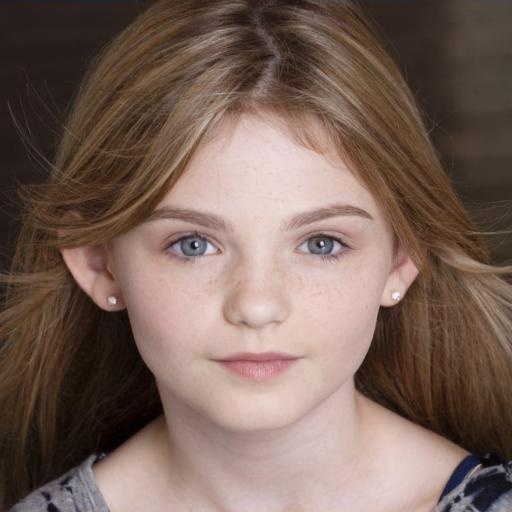}&
			\includegraphics[width=\widthj\linewidth]{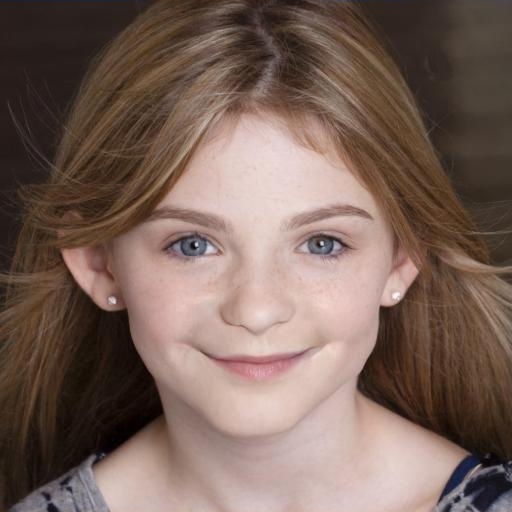}\\
			
			\includegraphics[width=\widthj\linewidth]{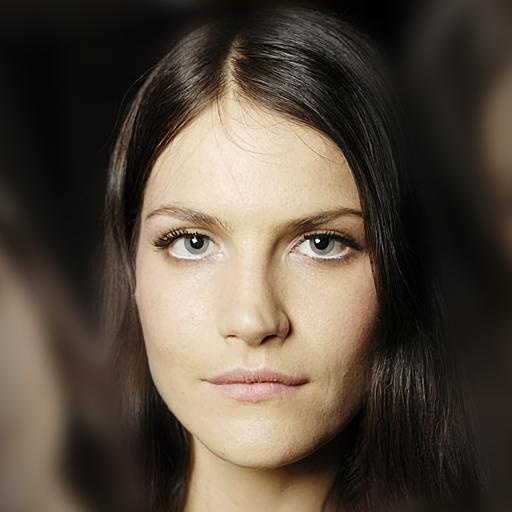}&
			\includegraphics[width=\widthj\linewidth]{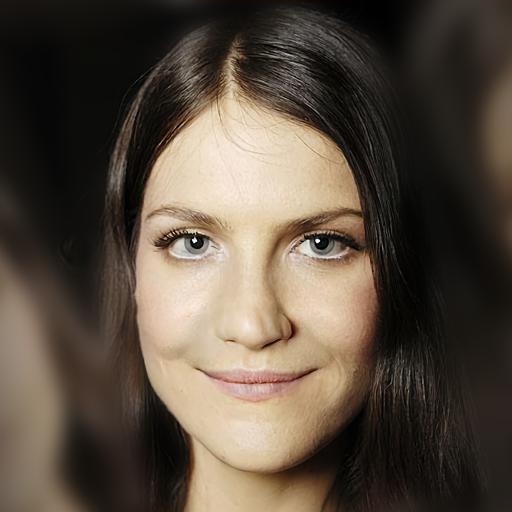}&
			
			\includegraphics[width=\widthj\linewidth]{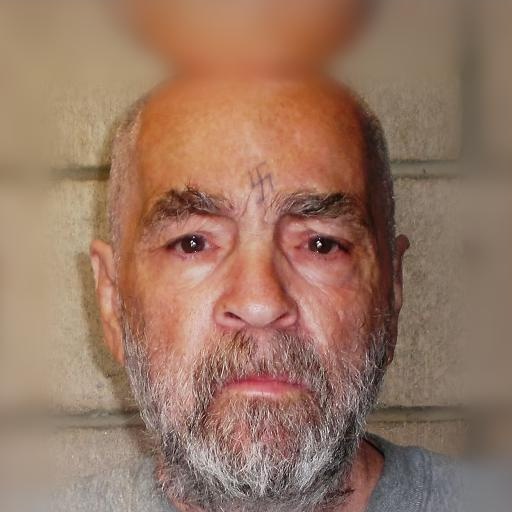}&
			\includegraphics[width=\widthj\linewidth]{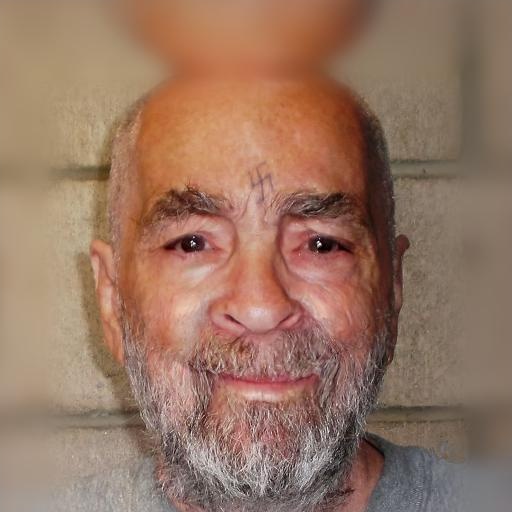}\\
			
			\includegraphics[width=\widthj\linewidth]{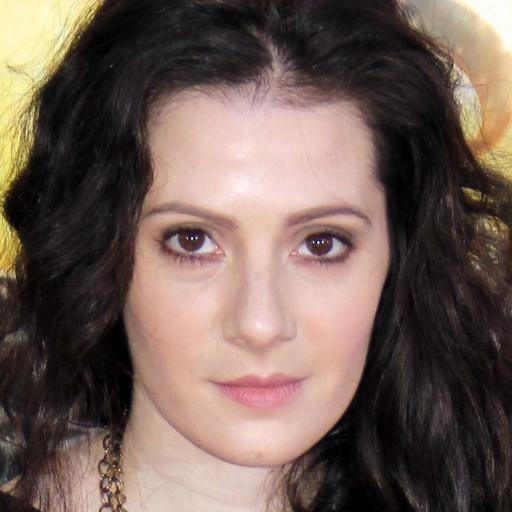}&
			\includegraphics[width=\widthj\linewidth]{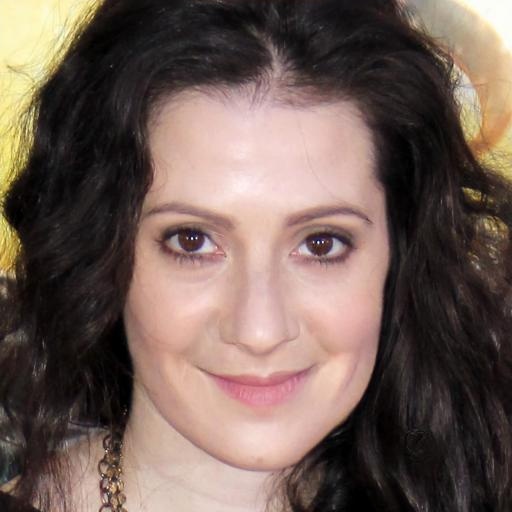}&
			
			\includegraphics[width=\widthj\linewidth]{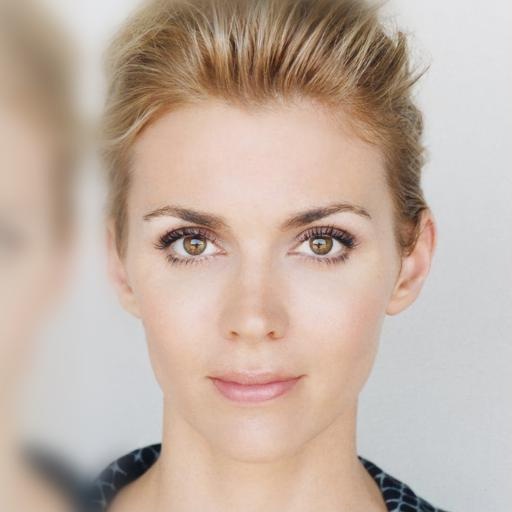}&
			\includegraphics[width=\widthj\linewidth]{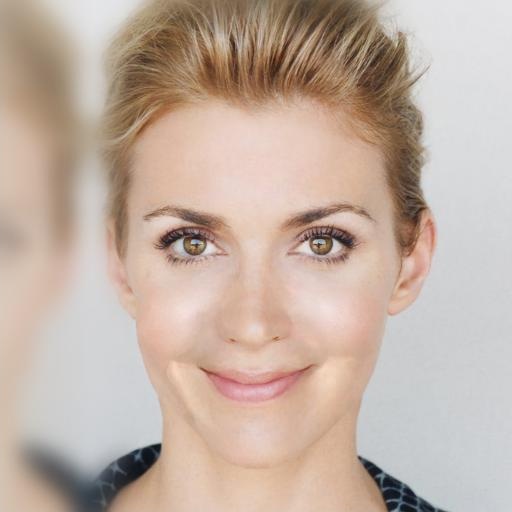}\\
			
			\includegraphics[width=\widthj\linewidth]{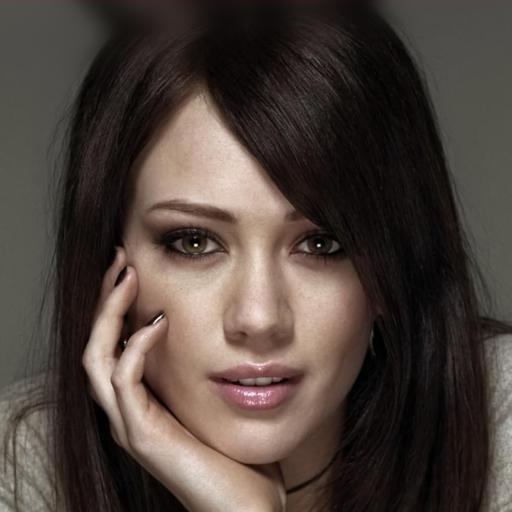}&
			\includegraphics[width=\widthj\linewidth]{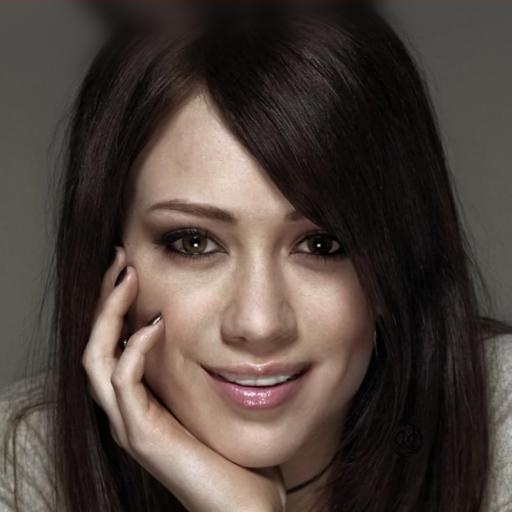}&
			
			\includegraphics[width=\widthj\linewidth]{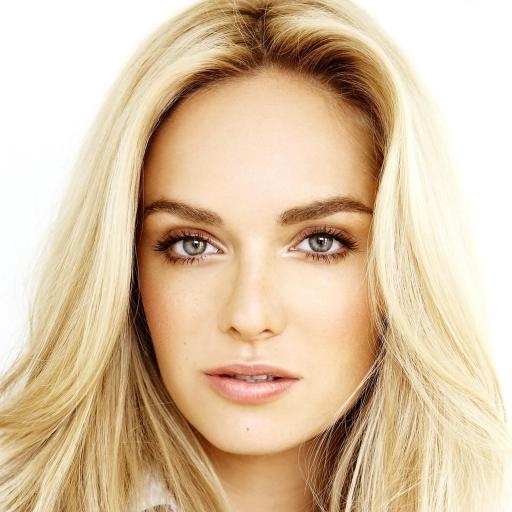}&
			\includegraphics[width=\widthj\linewidth]{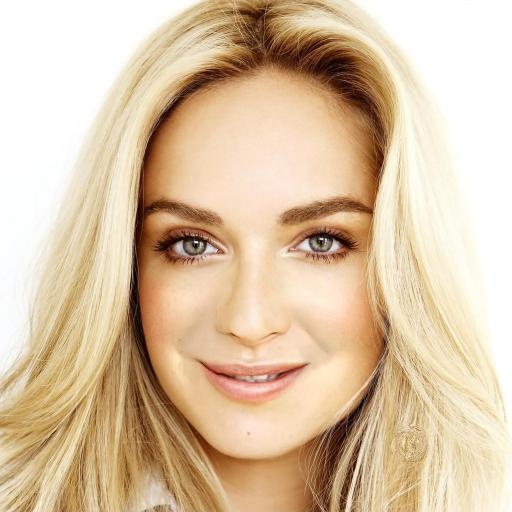}\\
			
			Input & $SPM + R$ & Input & $SPM + R$ \\
			
		\end{tabular}
		
	\end{center}
	\caption{Visualization of $512 \times 512$ results.}
	\label{fig:hires512}
\end{figure*}

\subsubsection{More Combination Cases}
\vspace{-0.05in}
More cases for combination of spontaneous motion basis are shown in Fig. \ref{fig:flow_comb}. It produces the combination of different facial attributes. 

\def\widthb{0.14}
\begin{figure*}
	\begin{center}
		\begin{tabular}
			{@{\hspace{0.0mm}}c@{\hspace{2mm}}c@{\hspace{1mm}}c@{\hspace{1mm}}c@{\hspace{1mm}}c@{\hspace{1mm}}c@{\hspace{1mm}}c@{\hspace{0.0mm}}}
			&
			\includegraphics[width=\widthb\linewidth]{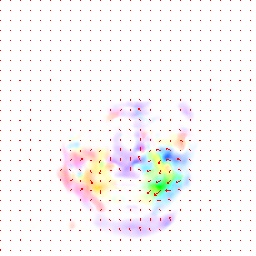}&
			\includegraphics[width=\widthb\linewidth]{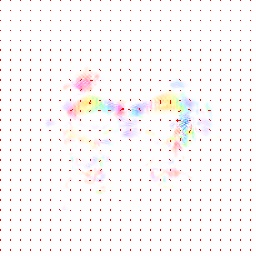}&
			\includegraphics[width=\widthb\linewidth]{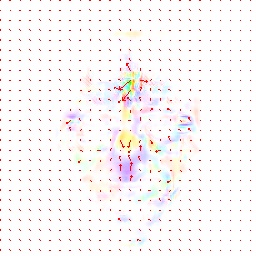}&
			\includegraphics[width=\widthb\linewidth]{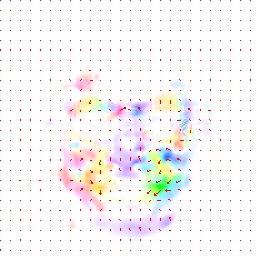} &
			\includegraphics[width=\widthb\linewidth]{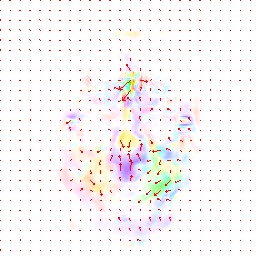} &
			\includegraphics[width=\widthb\linewidth]{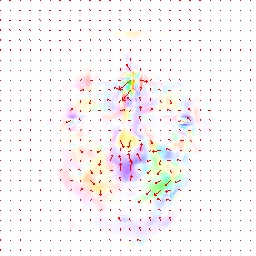} \\
			
			\includegraphics[width=\widthb\linewidth]{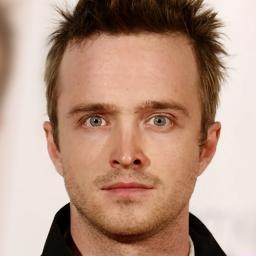}&
			\includegraphics[width=\widthb\linewidth]{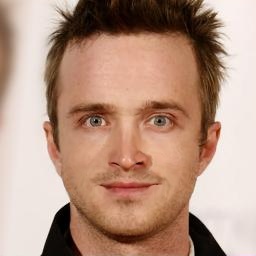}&
			\includegraphics[width=\widthb\linewidth]{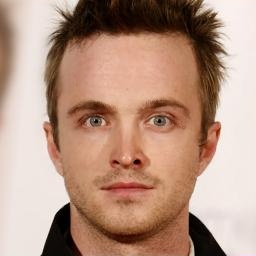}&
			\includegraphics[width=\widthb\linewidth]{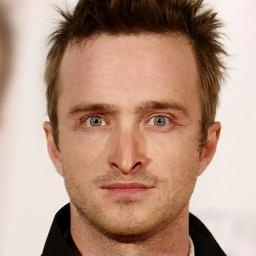}&
			\includegraphics[width=\widthb\linewidth]{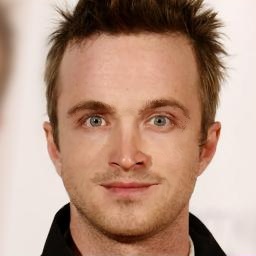} &
			\includegraphics[width=\widthb\linewidth]{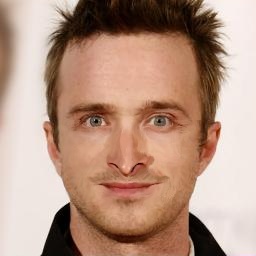} &
			\includegraphics[width=\widthb\linewidth]{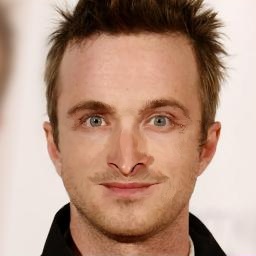} \\
			
			&
			\includegraphics[width=\widthb\linewidth]{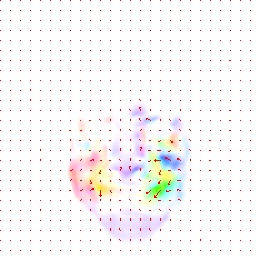}&
			\includegraphics[width=\widthb\linewidth]{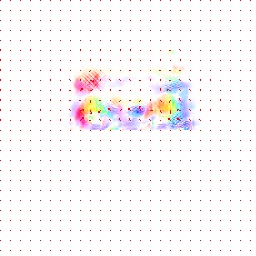}&
			\includegraphics[width=\widthb\linewidth]{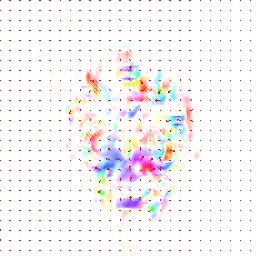}&
			\includegraphics[width=\widthb\linewidth]{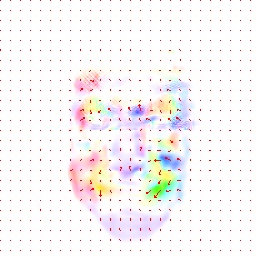} &
			\includegraphics[width=\widthb\linewidth]{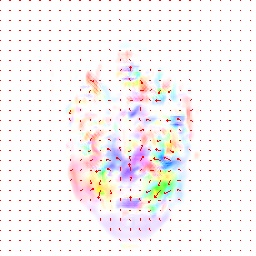} &
			\includegraphics[width=\widthb\linewidth]{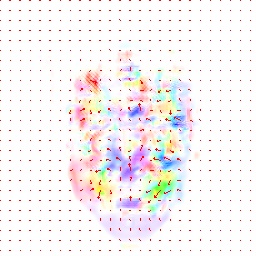} \\
			
			\includegraphics[width=\widthb\linewidth]{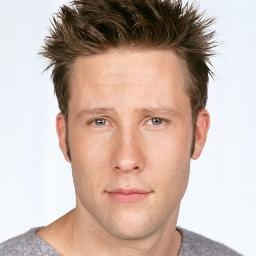}&
			\includegraphics[width=\widthb\linewidth]{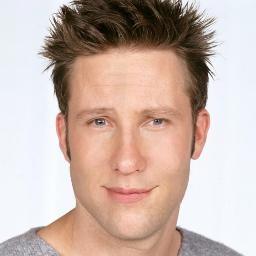}&
			\includegraphics[width=\widthb\linewidth]{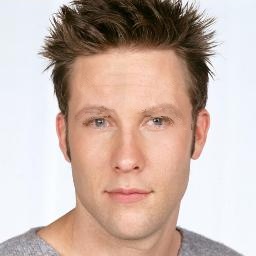}&
			\includegraphics[width=\widthb\linewidth]{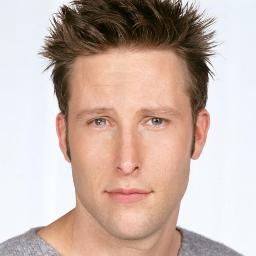}&
			\includegraphics[width=\widthb\linewidth]{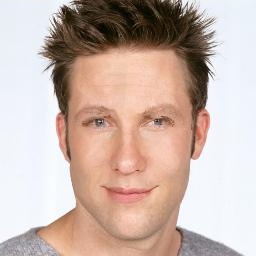} &
			\includegraphics[width=\widthb\linewidth]{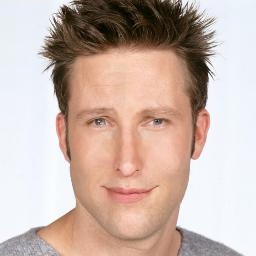} &
			\includegraphics[width=\widthb\linewidth]{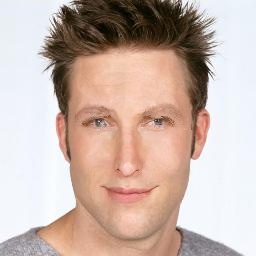} \\
			
			&
			\includegraphics[width=\widthb\linewidth]{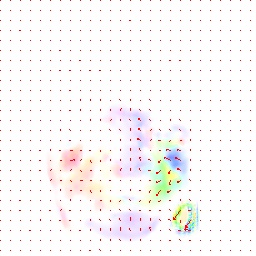}&
			\includegraphics[width=\widthb\linewidth]{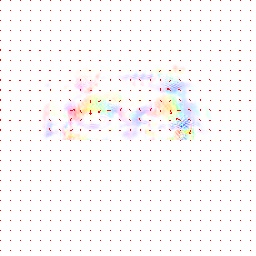}&
			\includegraphics[width=\widthb\linewidth]{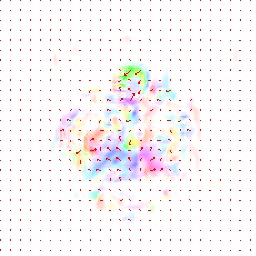}&
			\includegraphics[width=\widthb\linewidth]{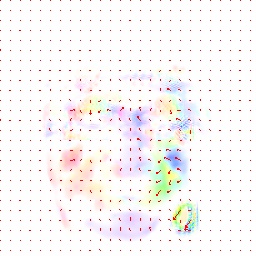} &
			\includegraphics[width=\widthb\linewidth]{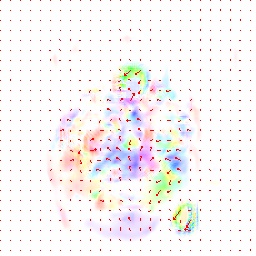} &
			\includegraphics[width=\widthb\linewidth]{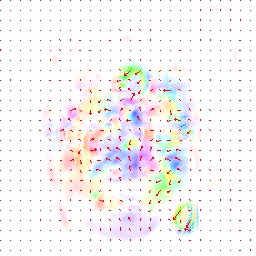} \\
			
			\includegraphics[width=\widthb\linewidth]{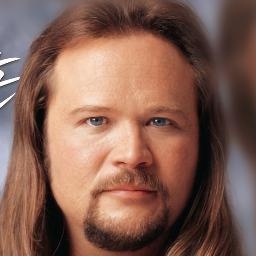}&
			\includegraphics[width=\widthb\linewidth]{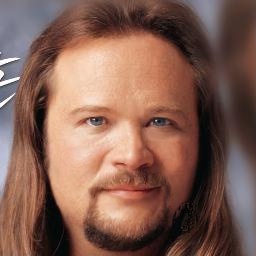}&
			\includegraphics[width=\widthb\linewidth]{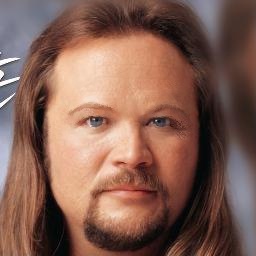}&
			\includegraphics[width=\widthb\linewidth]{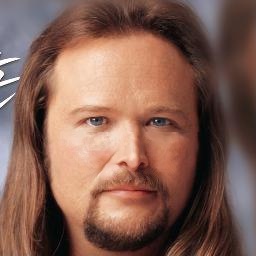}&
			\includegraphics[width=\widthb\linewidth]{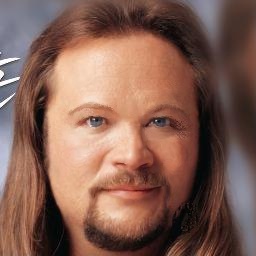} &
			\includegraphics[width=\widthb\linewidth]{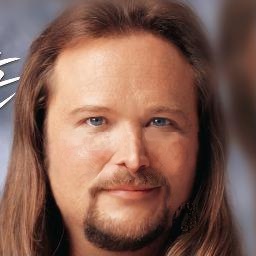} &
			\includegraphics[width=\widthb\linewidth]{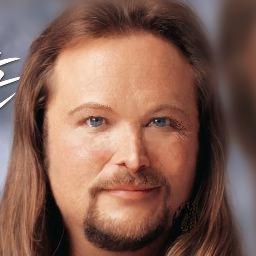} \\

			&
			\includegraphics[width=\widthb\linewidth]{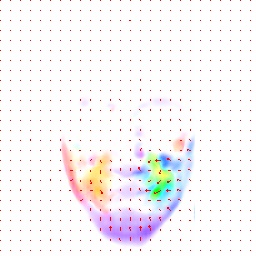}&
			\includegraphics[width=\widthb\linewidth]{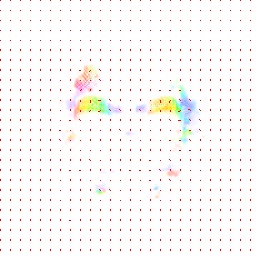}&
			\includegraphics[width=\widthb\linewidth]{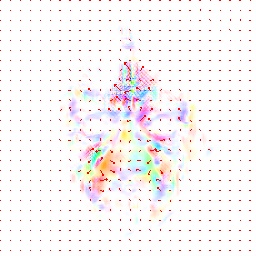}&
			\includegraphics[width=\widthb\linewidth]{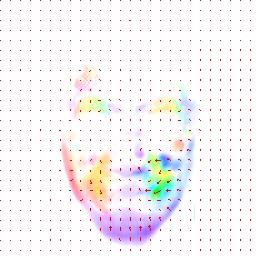} &
			\includegraphics[width=\widthb\linewidth]{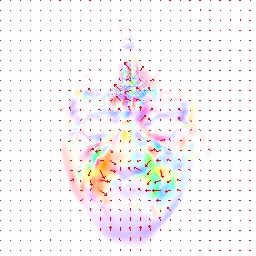} &
			\includegraphics[width=\widthb\linewidth]{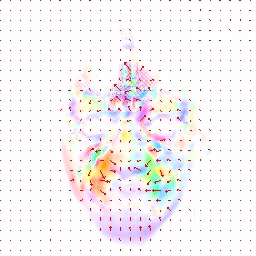} \\
			
			\includegraphics[width=\widthb\linewidth]{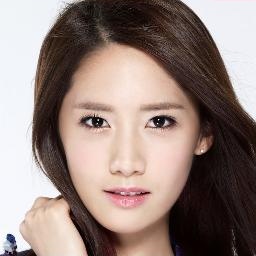}&
			\includegraphics[width=\widthb\linewidth]{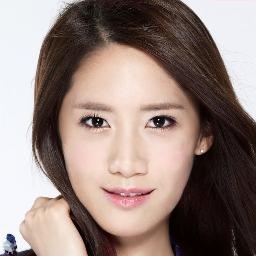}&
			\includegraphics[width=\widthb\linewidth]{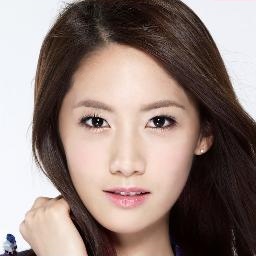}&
			\includegraphics[width=\widthb\linewidth]{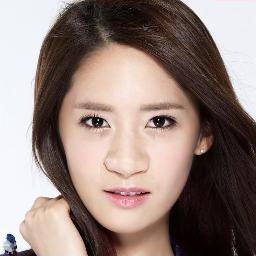}&
			\includegraphics[width=\widthb\linewidth]{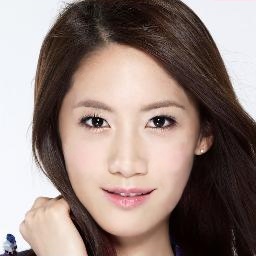} &
			\includegraphics[width=\widthb\linewidth]{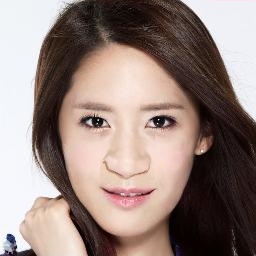} &
			\includegraphics[width=\widthb\linewidth]{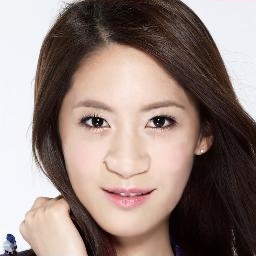} \\
			

			Input & $M_s$ & $M_e$ & $M_n$ & $M_s+M_e$ & $M_s+M_n$ & $M_s+M_e+M_n$ \\
			
		\end{tabular}
		
	\end{center}
	\vspace{-0.12in}
	\caption{More cases for motion field basis combination. For each case, first row: motion fields under different conditions, Second row: deformed results by applying corresponding motion fields. $M_s$: `Smiling' transform, $M_e$: `Arched\_Eyebrow' transform, $M_n$: `Pointy\_Nose' or `Big\_Nose' transform. $M_s+M_e$, $M_s+M_n$, $M_s+M_e+M_n$ are with two or three corresponding motion field combination. }
	\label{fig:flow_comb}
\end{figure*}

\subsection{More Visual Results}
\vspace{-0.05in}
More visual results with different facial attribute transforms are shown in Fig.~\ref{fig:visual} (CelebA dataset) and Fig.~\ref{fig:rafd} (RaFD dataset). 

\def\widthj{0.122}
\begin{figure*}[h]
	\begin{center}
		\begin{tabular}
			{@{\hspace{0.0mm}}c@{\hspace{0.5mm}}c@{\hspace{2mm}}c@{\hspace{0.5mm}}c@{\hspace{2mm}}c@{\hspace{0.5mm}}c@{\hspace{2mm}}c@{\hspace{0.5mm}}c@{\hspace{0.0mm}}}
			
			\includegraphics[width=\widthj\linewidth]{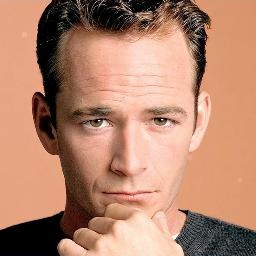}&
			\includegraphics[width=\widthj\linewidth]{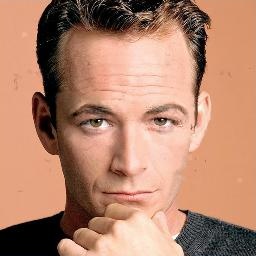}&
			
			\includegraphics[width=\widthj\linewidth]{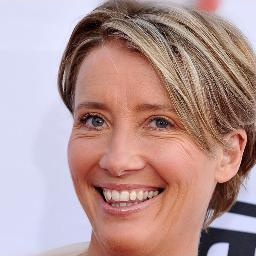}&
			\includegraphics[width=\widthj\linewidth]{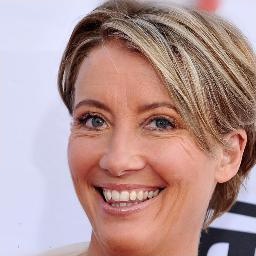}&
			
			\includegraphics[width=\widthj\linewidth]{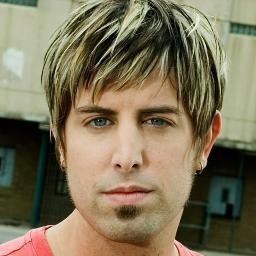}&
			\includegraphics[width=\widthj\linewidth]{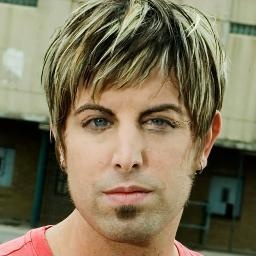}&
			
			\includegraphics[width=\widthj\linewidth]{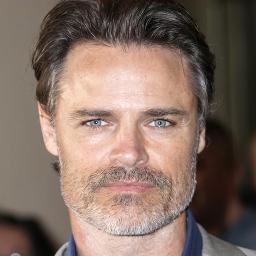}&
			\includegraphics[width=\widthj\linewidth]{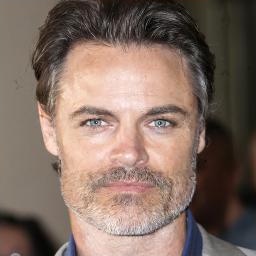}\\
			
			\includegraphics[width=\widthj\linewidth]{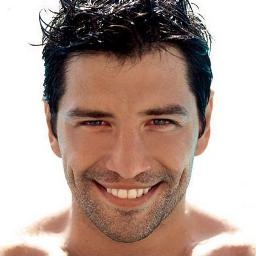}&
			\includegraphics[width=\widthj\linewidth]{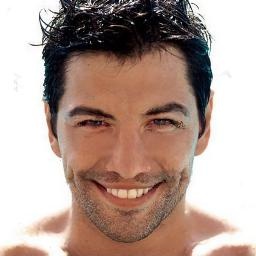}&
			
			\includegraphics[width=\widthj\linewidth]{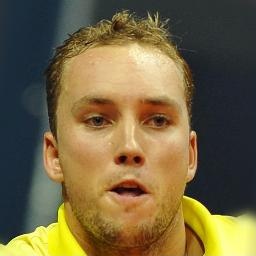}&
			\includegraphics[width=\widthj\linewidth]{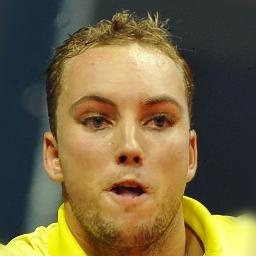}&
			
			\includegraphics[width=\widthj\linewidth]{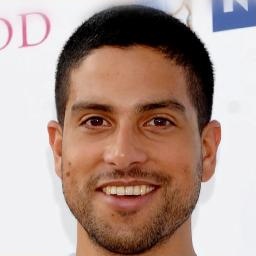}&
			\includegraphics[width=\widthj\linewidth]{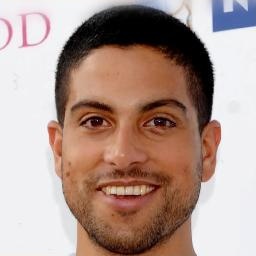}&
			
			\includegraphics[width=\widthj\linewidth]{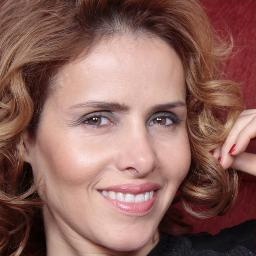}&
			\includegraphics[width=\widthj\linewidth]{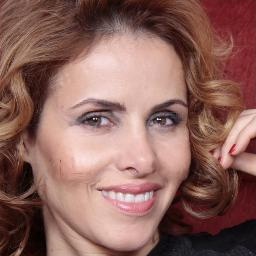}\\
			
			\includegraphics[width=\widthj\linewidth]{figs/fig_eyebrow/inp_21_8.jpg}&
			\includegraphics[width=\widthj\linewidth]{figs/fig_eyebrow/ours_21_8.jpg}&
			
			\includegraphics[width=\widthj\linewidth]{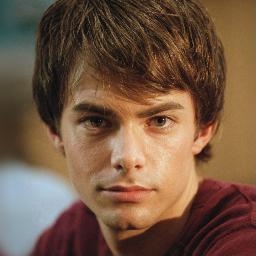}&
			\includegraphics[width=\widthj\linewidth]{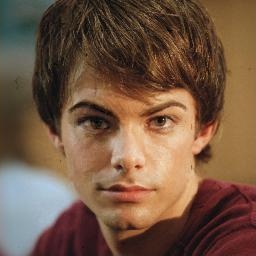}&
			
			\includegraphics[width=\widthj\linewidth]{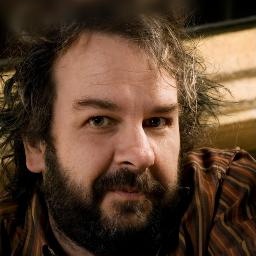}&
			\includegraphics[width=\widthj\linewidth]{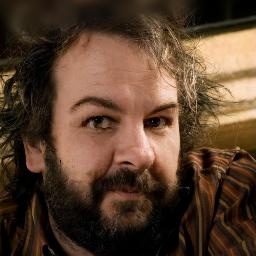}&
			
			\includegraphics[width=\widthj\linewidth]{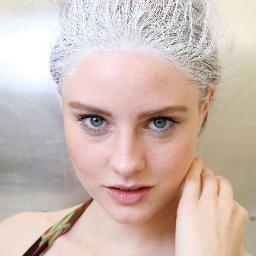}&
			\includegraphics[width=\widthj\linewidth]{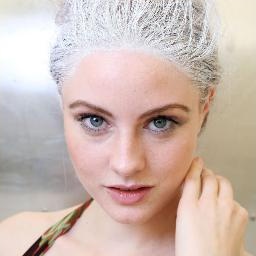}\\
			
		\end{tabular}
	\end{center}
	
	\begin{center}
		\vspace{-0.15in}
		(a) Arched\_Eyebrow.
		\vspace{-0.15in}
	\end{center}
	
	\begin{center}
		\vspace{-0.1in}
		\begin{tabular}	
			{@{\hspace{0.0mm}}c@{\hspace{0.5mm}}c@{\hspace{2mm}}c@{\hspace{0.5mm}}c@{\hspace{2mm}}c@{\hspace{0.5mm}}c@{\hspace{2mm}}c@{\hspace{0.5mm}}c@{\hspace{0.0mm}}}	
			\includegraphics[width=\widthj\linewidth]{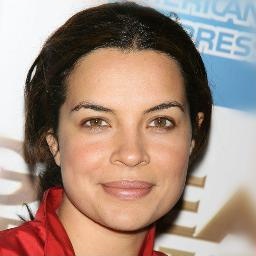}&
			\includegraphics[width=\widthj\linewidth]{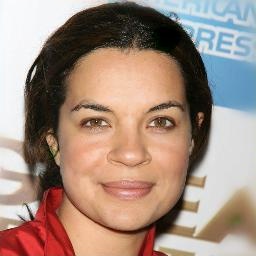}&
			
			\includegraphics[width=\widthj\linewidth]{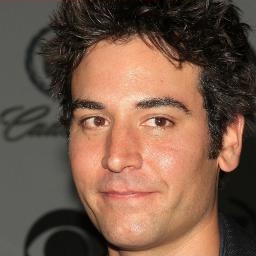}&
			\includegraphics[width=\widthj\linewidth]{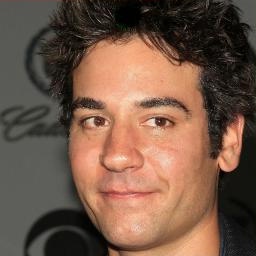}&
			
			\includegraphics[width=\widthj\linewidth]{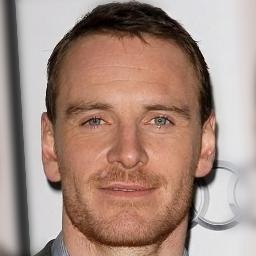}&
			\includegraphics[width=\widthj\linewidth]{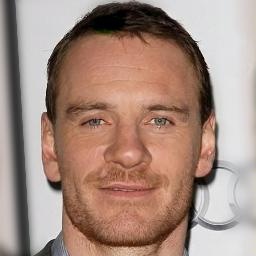}&
			
			\includegraphics[width=\widthj\linewidth]{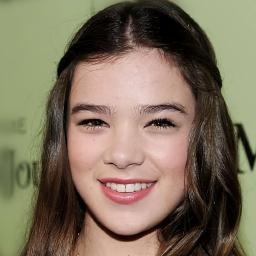}&
			\includegraphics[width=\widthj\linewidth]{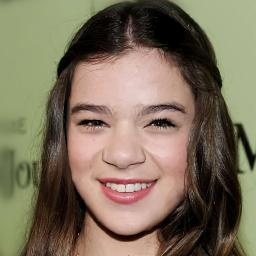}\\
			
			\includegraphics[width=\widthj\linewidth]{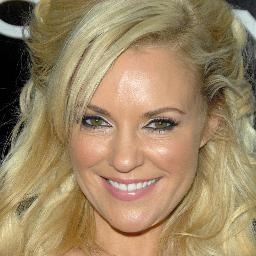}&
			\includegraphics[width=\widthj\linewidth]{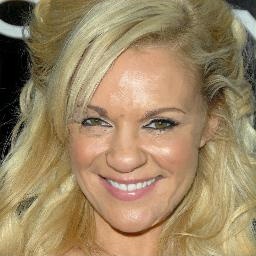}&
			
			\includegraphics[width=\widthj\linewidth]{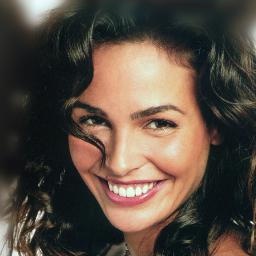}&
			\includegraphics[width=\widthj\linewidth]{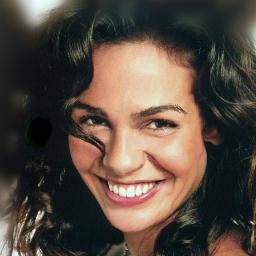}&
			
			\includegraphics[width=\widthj\linewidth]{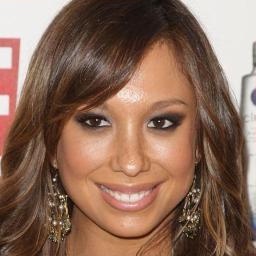}&
			\includegraphics[width=\widthj\linewidth]{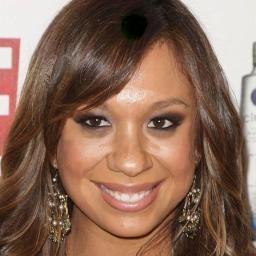}&
			
			\includegraphics[width=\widthj\linewidth]{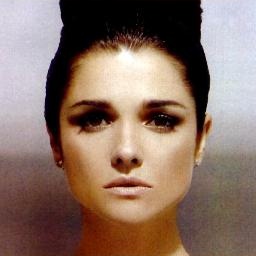}&
			\includegraphics[width=\widthj\linewidth]{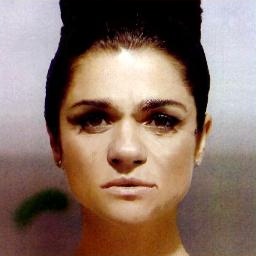}\\
			
			\includegraphics[width=\widthj\linewidth]{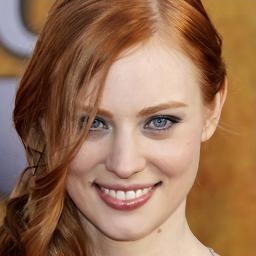}&
			\includegraphics[width=\widthj\linewidth]{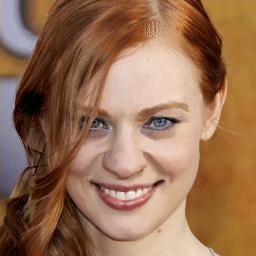}&
			
			\includegraphics[width=\widthj\linewidth]{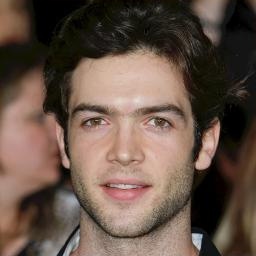}&
			\includegraphics[width=\widthj\linewidth]{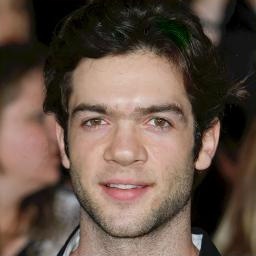}&
			
			\includegraphics[width=\widthj\linewidth]{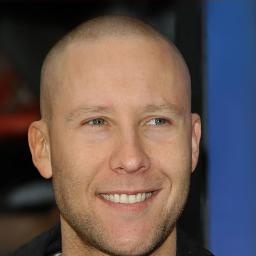}&
			\includegraphics[width=\widthj\linewidth]{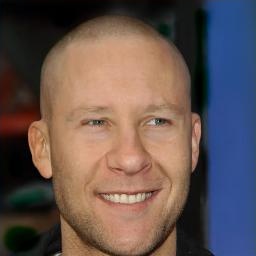}&
			
			\includegraphics[width=\widthj\linewidth]{figs/fig_big_nose/inp_10_11.jpg}&
			\includegraphics[width=\widthj\linewidth]{figs/fig_big_nose/ours_10_11.jpg}\\
		\end{tabular}
	\end{center}
	\begin{center}
		\vspace{-0.15in}
		(b) Big\_Nose.
		\vspace{-0.15in}
	\end{center}
	
	\begin{center}
		\vspace{-0.1in}
		\begin{tabular}		
			{@{\hspace{0.0mm}}c@{\hspace{0.5mm}}c@{\hspace{2mm}}c@{\hspace{0.5mm}}c@{\hspace{2mm}}c@{\hspace{0.5mm}}c@{\hspace{2mm}}c@{\hspace{0.5mm}}c@{\hspace{0.0mm}}}
			\includegraphics[width=\widthj\linewidth]{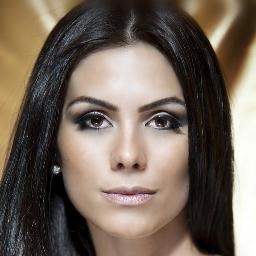}&
			\includegraphics[width=\widthj\linewidth]{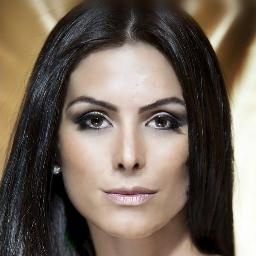}&
			
			\includegraphics[width=\widthj\linewidth]{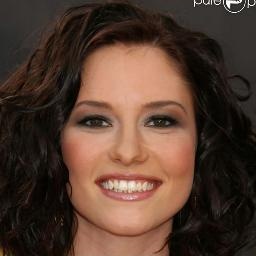}&
			\includegraphics[width=\widthj\linewidth]{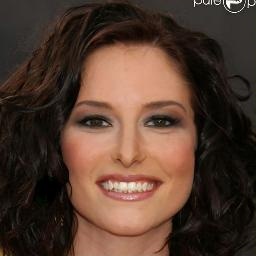}&
			
			\includegraphics[width=\widthj\linewidth]{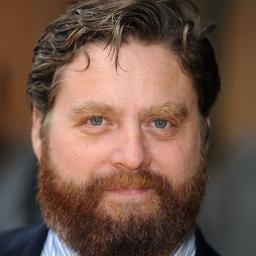}&
			\includegraphics[width=\widthj\linewidth]{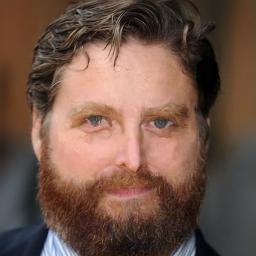}&
			
			\includegraphics[width=\widthj\linewidth]{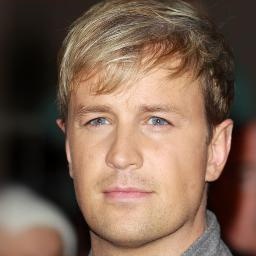}&
			\includegraphics[width=\widthj\linewidth]{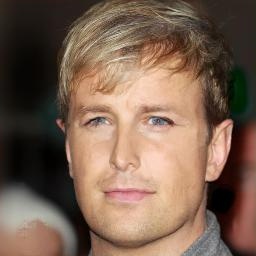}\\
			
			\includegraphics[width=\widthj\linewidth]{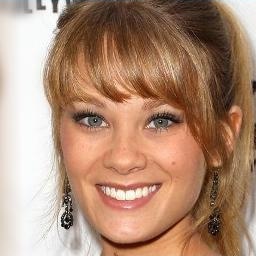}&
			\includegraphics[width=\widthj\linewidth]{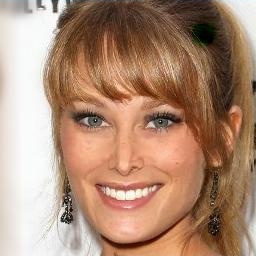}&
			
			\includegraphics[width=\widthj\linewidth]{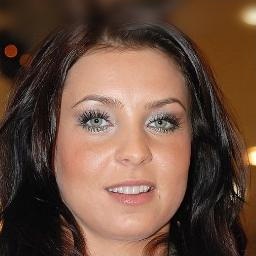}&
			\includegraphics[width=\widthj\linewidth]{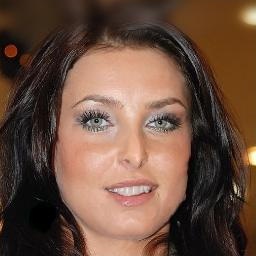}&
			
			\includegraphics[width=\widthj\linewidth]{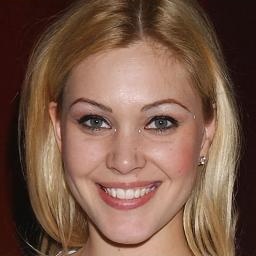}&
			\includegraphics[width=\widthj\linewidth]{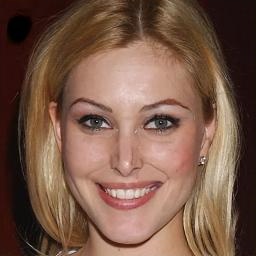}&
			
			\includegraphics[width=\widthj\linewidth]{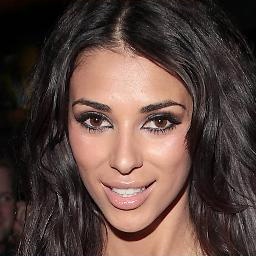}&
			\includegraphics[width=\widthj\linewidth]{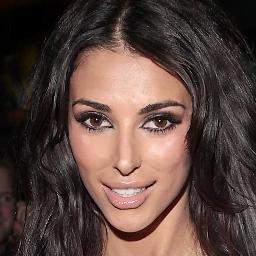}\\
			
			\includegraphics[width=\widthj\linewidth]{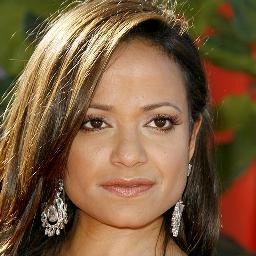}&
			\includegraphics[width=\widthj\linewidth]{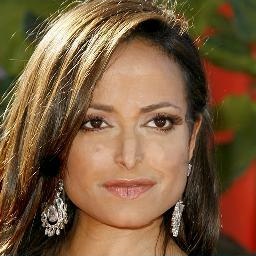}&
			
			\includegraphics[width=\widthj\linewidth]{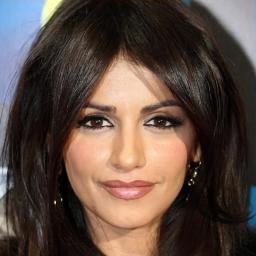}&
			\includegraphics[width=\widthj\linewidth]{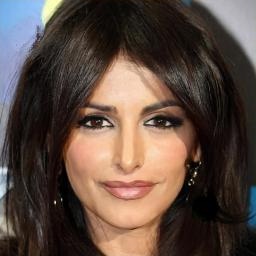}&
			
			\includegraphics[width=\widthj\linewidth]{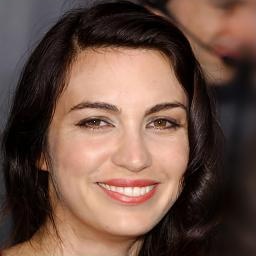}&
			\includegraphics[width=\widthj\linewidth]{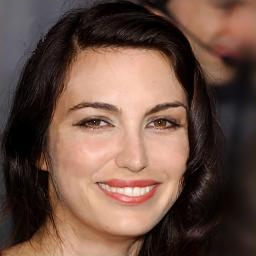}&
			
			\includegraphics[width=\widthj\linewidth]{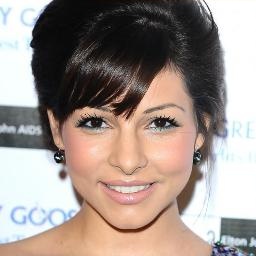}&
			\includegraphics[width=\widthj\linewidth]{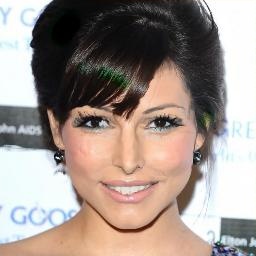}\\
			
			
		\end{tabular}
	\end{center}
	\begin{center}
		\vspace{-0.15in}
		(c) Pointy\_Nose.
	\end{center}
	
	\caption{More visual results on CelebA dataset with different attribute transformations: (a) ``Arched\_Eyebrow'', (b) ``Big\_Nose'' and (c) ``Pointy\_Nose''. For each example, left: Input image, right: Our generated result.}
	\label{fig:visual}
\end{figure*}

\def\widthm{0.117}
\begin{figure*}
	\begin{center}
		\begin{tabular}
			{@{\hspace{0.0mm}}c@{\hspace{3mm}}c@{\hspace{1.0mm}}c@{\hspace{1.0mm}}c@{\hspace{1.0mm}}c@{\hspace{1.0mm}}c@{\hspace{1.0mm}}c@{\hspace{1.0mm}}c@{\hspace{0mm}}}
			\includegraphics[width=\widthm\linewidth]{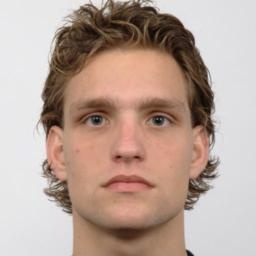} &
			\includegraphics[width=\widthm\linewidth]{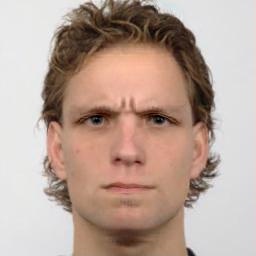}&
			\includegraphics[width=\widthm\linewidth]{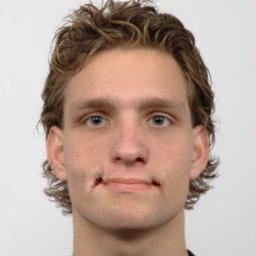}&
			\includegraphics[width=\widthm\linewidth]{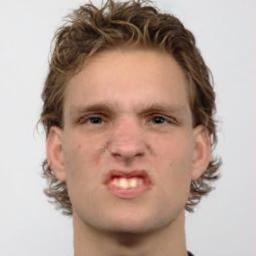}&
			\includegraphics[width=\widthm\linewidth]{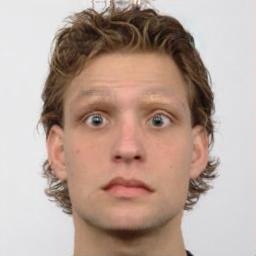}&
			\includegraphics[width=\widthm\linewidth]{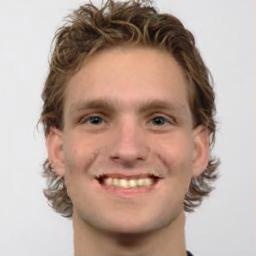}&
			\includegraphics[width=\widthm\linewidth]{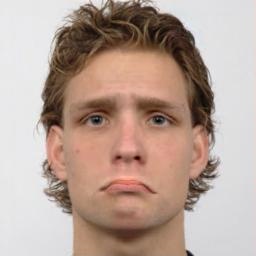}&
			\includegraphics[width=\widthm\linewidth]{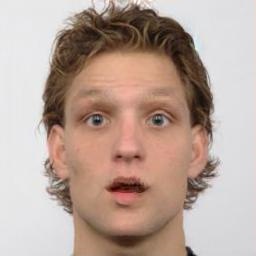}\\
			
			\includegraphics[width=\widthm\linewidth]{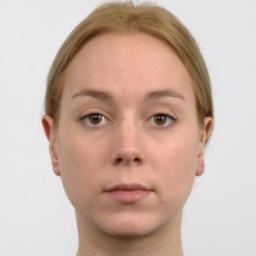} &
			\includegraphics[width=\widthm\linewidth]{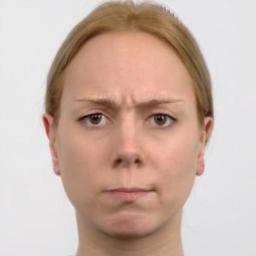}&
			\includegraphics[width=\widthm\linewidth]{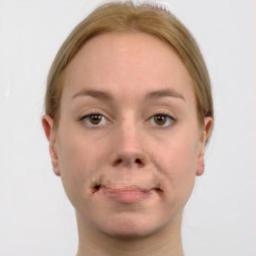}&
			\includegraphics[width=\widthm\linewidth]{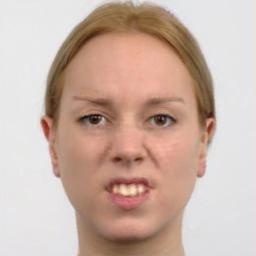}&
			\includegraphics[width=\widthm\linewidth]{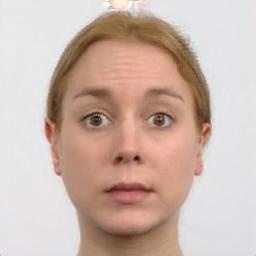}&
			\includegraphics[width=\widthm\linewidth]{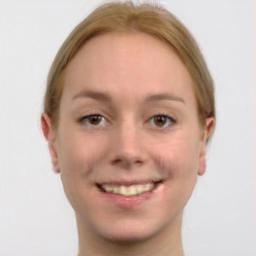}&
			\includegraphics[width=\widthm\linewidth]{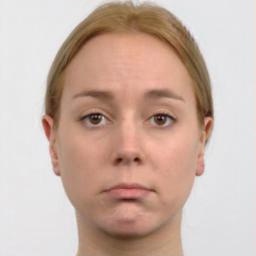}&
			\includegraphics[width=\widthm\linewidth]{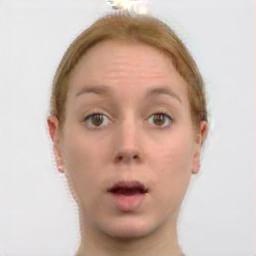}\\
			
			\includegraphics[width=\widthm\linewidth]{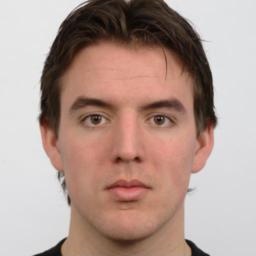} &
			\includegraphics[width=\widthm\linewidth]{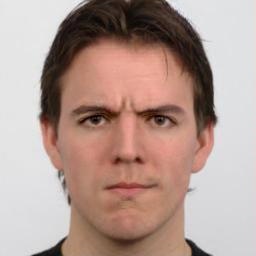}&
			\includegraphics[width=\widthm\linewidth]{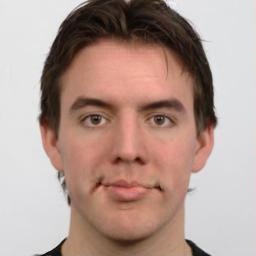}&
			\includegraphics[width=\widthm\linewidth]{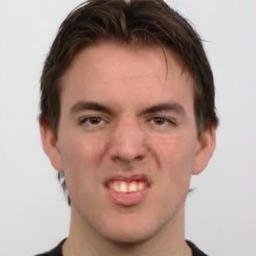}&
			\includegraphics[width=\widthm\linewidth]{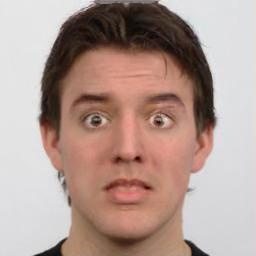}&
			\includegraphics[width=\widthm\linewidth]{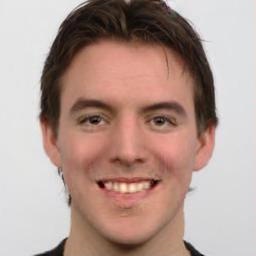}&
			\includegraphics[width=\widthm\linewidth]{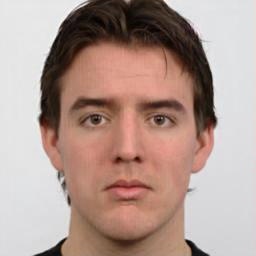}&
			\includegraphics[width=\widthm\linewidth]{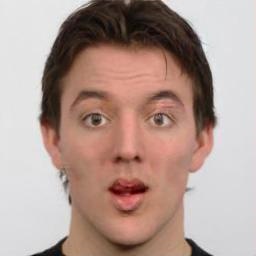}\\
			
			\includegraphics[width=\widthm\linewidth]{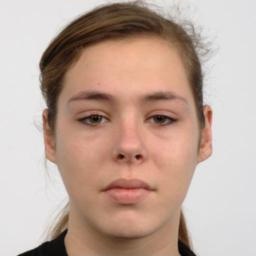} &
			\includegraphics[width=\widthm\linewidth]{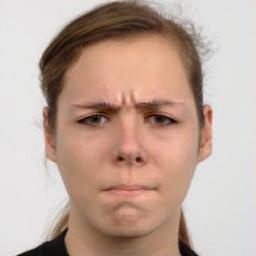}&
			\includegraphics[width=\widthm\linewidth]{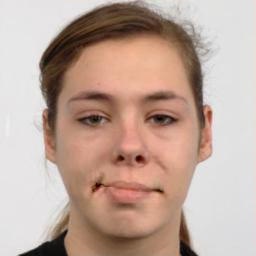}&
			\includegraphics[width=\widthm\linewidth]{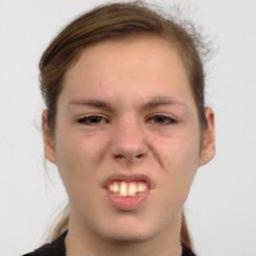}&
			\includegraphics[width=\widthm\linewidth]{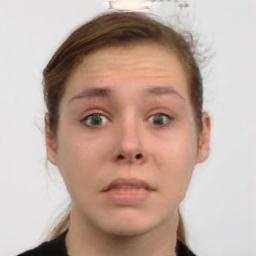}&
			\includegraphics[width=\widthm\linewidth]{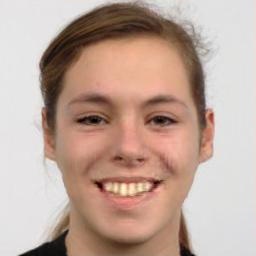}&
			\includegraphics[width=\widthm\linewidth]{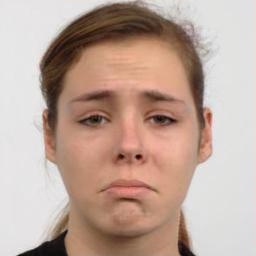}&
			\includegraphics[width=\widthm\linewidth]{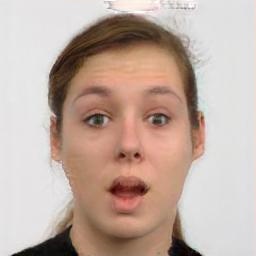}\\

			\includegraphics[width=\widthm\linewidth]{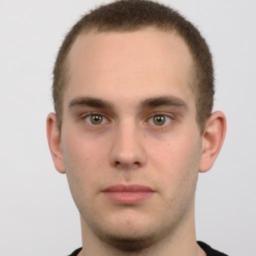} &
			\includegraphics[width=\widthm\linewidth]{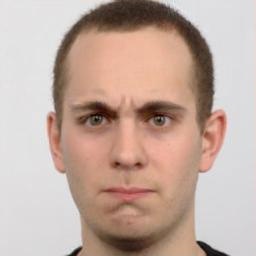}&
			\includegraphics[width=\widthm\linewidth]{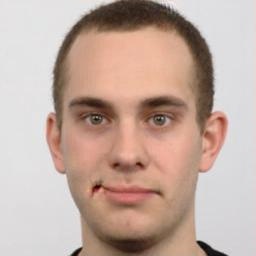}&
			\includegraphics[width=\widthm\linewidth]{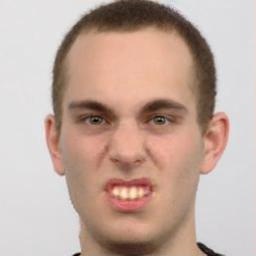}&
			\includegraphics[width=\widthm\linewidth]{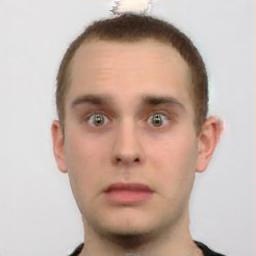}&
			\includegraphics[width=\widthm\linewidth]{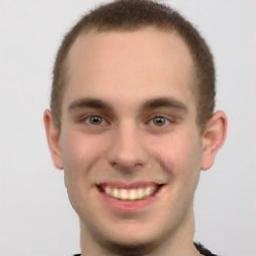}&
			\includegraphics[width=\widthm\linewidth]{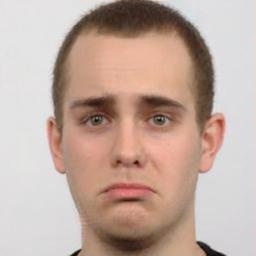}&
			\includegraphics[width=\widthm\linewidth]{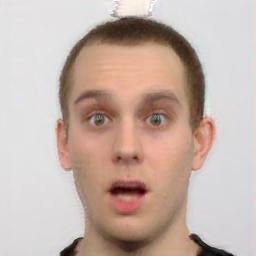}\\

			\includegraphics[width=\widthm\linewidth]{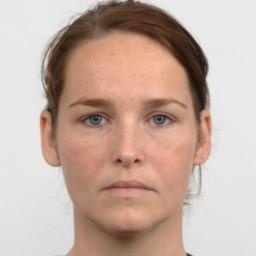} &
			\includegraphics[width=\widthm\linewidth]{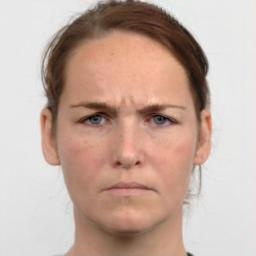}&
			\includegraphics[width=\widthm\linewidth]{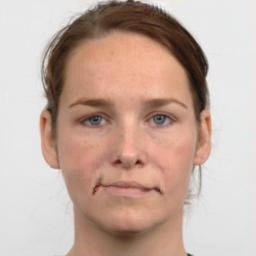}&
			\includegraphics[width=\widthm\linewidth]{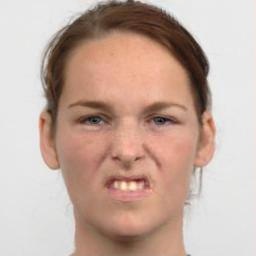}&
			\includegraphics[width=\widthm\linewidth]{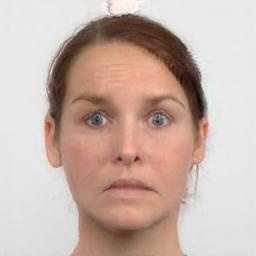}&
			\includegraphics[width=\widthm\linewidth]{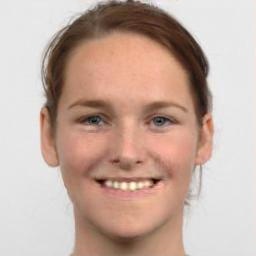}&
			\includegraphics[width=\widthm\linewidth]{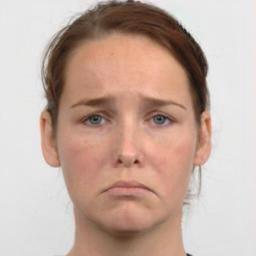}&
			\includegraphics[width=\widthm\linewidth]{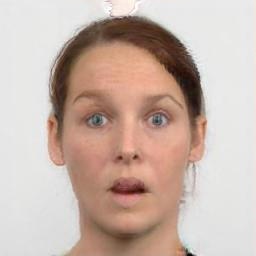}\\

			\includegraphics[width=\widthm\linewidth]{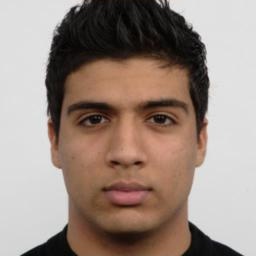} &
			\includegraphics[width=\widthm\linewidth]{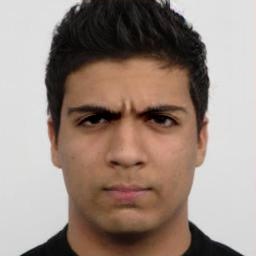}&
			\includegraphics[width=\widthm\linewidth]{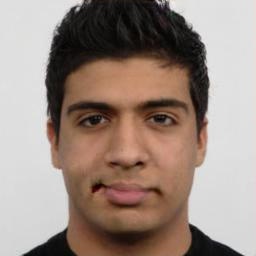}&
			\includegraphics[width=\widthm\linewidth]{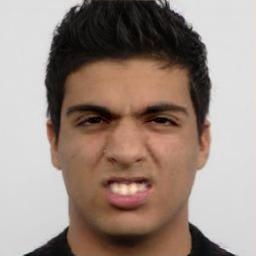}&
			\includegraphics[width=\widthm\linewidth]{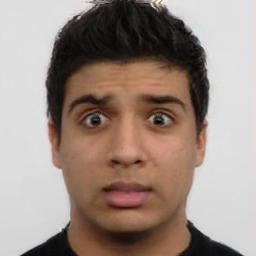}&
			\includegraphics[width=\widthm\linewidth]{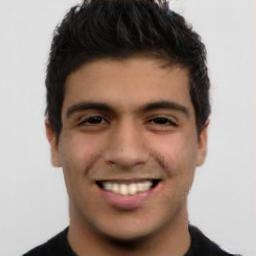}&
			\includegraphics[width=\widthm\linewidth]{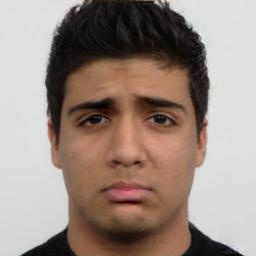}&
			\includegraphics[width=\widthm\linewidth]{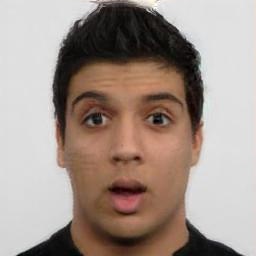}\\
			
			\includegraphics[width=\widthm\linewidth]{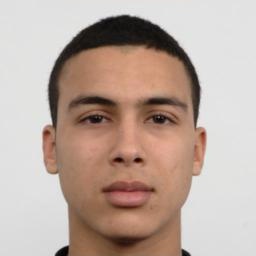} &
			\includegraphics[width=\widthm\linewidth]{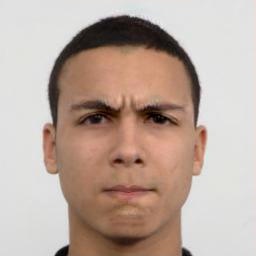}&
			\includegraphics[width=\widthm\linewidth]{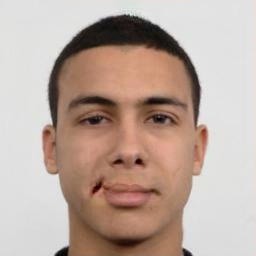}&
			\includegraphics[width=\widthm\linewidth]{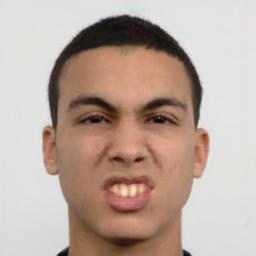}&
			\includegraphics[width=\widthm\linewidth]{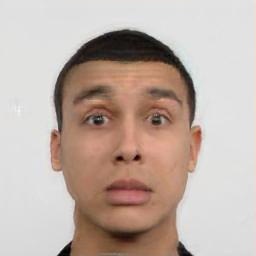}&
			\includegraphics[width=\widthm\linewidth]{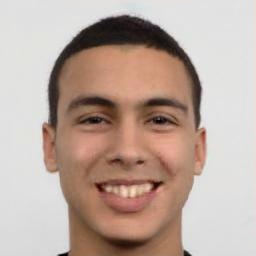}&
			\includegraphics[width=\widthm\linewidth]{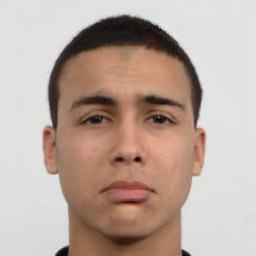}&
			\includegraphics[width=\widthm\linewidth]{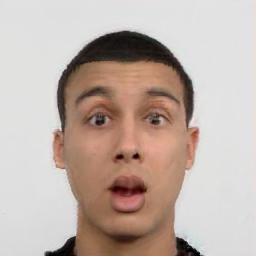}\\
			
			\includegraphics[width=\widthm\linewidth]{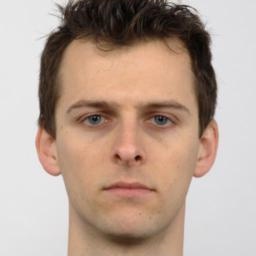} &
			\includegraphics[width=\widthm\linewidth]{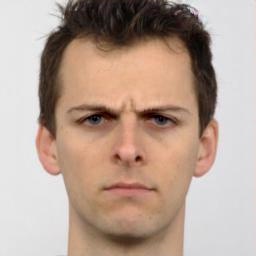}&
			\includegraphics[width=\widthm\linewidth]{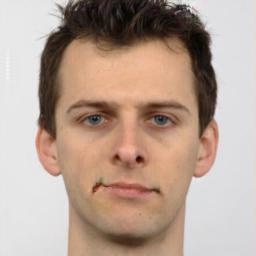}&
			\includegraphics[width=\widthm\linewidth]{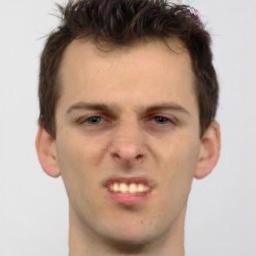}&
			\includegraphics[width=\widthm\linewidth]{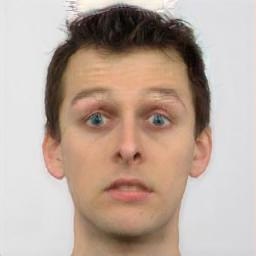}&
			\includegraphics[width=\widthm\linewidth]{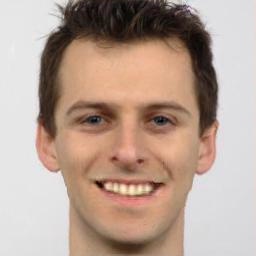}&
			\includegraphics[width=\widthm\linewidth]{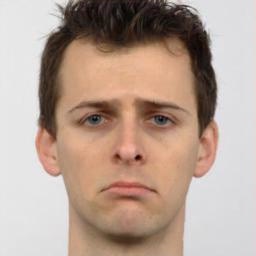}&
			\includegraphics[width=\widthm\linewidth]{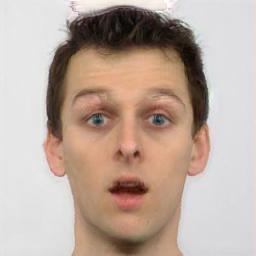}\\
			
			Input & Angry & Contemptuous & Disgusted & Fearful & Happy & Sad & Surprised \\
		\end{tabular}
	\end{center}
	\vspace{0.1in}
	\caption{More visual results on the RaFD dataset with different expression transformations.}
	\label{fig:rafd}
\end{figure*}

\subsection{Network Architecture}
We follow the design of \cite{choi2018stargan} to set up our encoder-decoder structure for different modules of generator. The architecture detail of each module is illustrated as below. 

For the layers demonstrated in Tab.~\ref{tab:SPM}, \ref{tab:AMM} \& \ref{tab:RM}, ``conv$i$'' and ``deconv$i$'' indicate a convolution layer and a de-convolution layer respectively, each followed by an instance-normalization and a ReLU layer. Specifically, ``conv\_output'' indicates a convolution layer without normalization or activation layer. Besides, ``ResBlock'' indicates residual blocks defined in \cite{He_2016_CVPR}, while the original batch-normalization layer is replaced by instance-normalization layer and $\times n$ means $n$ residual blocks are stacked.
\paragraph{Spontaneous Motion Module}
Spontaneous motion module ($\textbf{SPM}$) is an image-attribute pair to motion field network, we construct the module with an encoder-decoder framework. The detail of network architecture is shown in Tab. \ref{tab:SPM}. 
\begin{table}[h]
	\renewcommand\arraystretch{1.3}
	\centering
	\begin{tabular}
		{|c|c|c|}
		\hline
		Layer & Output Size & ~(kernel, stride)~\\
		
		\hline 
		Inputs & $H \times W \times (3 + N)$ & (- , -) \\
		\hline \hline
		conv1 & $H \times W \times 64$ & (7, 1) \\
		\hline
		conv2 & $\frac{H}{2} \times \frac{W}{2} \times 128$ & (4, 2) \\
		\hline
		conv3 & $\frac{H}{4} \times \frac{W}{4} \times 256$ & (4, 2) \\
		\hline
		ResBlock $\times 6$ & $\frac{H}{4} \times \frac{W}{4} \times 256$ & (3, 1) \\
		\hline
		deconv1 & $\frac{H}{2} \times \frac{W}{2} \times 128$ & (4, 2) \\
		\hline
		deconv2 & $H \times W \times 64$ & (4, 2) \\
		\hline
		conv\_output & $H \times W \times 3$ & (7, 1) \\
		\hline
		Tanh & $H \times W \times 3$ & (-, -) \\
		\hline
	\end{tabular}
	\vspace{0.1in}
	\caption{Spontaneous Motion Module architecture. $N$ indicates the number of target attributes, $H$ and $W$ indicate height and width of the input images respectively. }
	\label{tab:SPM}
\end{table}

\paragraph{Attention Mask Module}
Attention mask module ($\textbf{M}$) is utilized to focus on the essential part for refinement by predicting a mask in $[0, 1]$. The structure of $\textbf{M}$ is a decoder, which takes extracted feature after ResBlocks from $\textbf{SPM}$ as input and outputs a one-channel mask. The network detail of $\textbf{M}$ is shown in Tab. \ref{tab:AMM}.
\begin{table}[h]
	\renewcommand\arraystretch{1.3}
	\centering
	\begin{tabular}
		{|c|c|c|}
		\hline
		Layer & Output Size & ~(kernel, stride)~\\
		\hline
		Inputs &  $\frac{H}{4} \times \frac{W}{4} \times 256$ & (- , -) \\
		\hline
		\hline
		deconv1 & $\frac{H}{2} \times \frac{W}{2} \times 128$ & (4, 2) \\
		\hline
		deconv2 & $H \times W \times 64$ & (4, 2) \\
		\hline
		conv\_output & $H \times W \times 1$ & (7, 1) \\
		\hline
		Sigmoid & $H \times W \times 1$ & (-, -) \\
		\hline
	\end{tabular}
	\vspace{0.1in}
	\caption{Attention mask module architecture. $H$ and $W$ indicate height and width of the input image respectively.}
	\label{tab:AMM}
\end{table}

\paragraph{Refinement Module}
Refinement module $\textbf{R}$ takes deformed image from $\textbf{SPM}$ as input, and outputs the refined image by introducing $n$ residual blocks. $\textbf{R}$ takes no down-sampling operation in order to keep the spatial information and refine the fine structure of deformed image. The network detail of $\textbf{R}$ is shown in Tab. \ref{tab:RM}.
\begin{table}[h]
	\renewcommand\arraystretch{1.3}
	\centering
	\begin{tabular}
		{|c|c|c|}
		\hline
		Layer & Output Size & ~(kernel, stride)~\\
		\hline
		Inputs & $H \times W \times 3$ & (- , -) \\
		\hline
		\hline
		conv1 & $H \times W \times 64$ & (7, 1) \\
		\hline
		ResBlock $\times n$ & $H \times W \times 64$ & (3, 1) \\
		\hline
		conv\_output & $H \times W \times 3$ & (7, 1) \\
		\hline
		Tanh & $H \times W \times 3$ & (-, -) \\
		\hline
	\end{tabular}
	\vspace{0.1in}
	\caption{Refinement module architecture. $H$ and $W$ refer to height and width of the input image respectively, and $n$ is set empirically according to our experiments.}
	\label{tab:RM}
\end{table}

\end{document}